\documentclass[11pt, a4paper, logo, copyright, nonumbering]{map}

\usepackage{geometry}
\usepackage{multirow}
\usepackage[most]{tcolorbox}
\usepackage[justification=justified]{caption}
\usepackage{subcaption} 
\usepackage{multicol}

\definecolor{cvprblue}{rgb}{0.21,0.49,0.74}
\definecolor{MidnightBlue}{RGB}{25,25,112}

\usepackage{natbib}

\usepackage{CJKutf8}

\usepackage{xargs}  

\usepackage{todonotes}  

\usepackage{multirow}

\usepackage{cleveref}

\usepackage{amsmath}
\usepackage{dsfont}

\usepackage{subcaption}

\usepackage{svg}
\PassOptionsToPackage{most}{tcolorbox}  
\usepackage{tcolorbox}  
\usepackage{fontawesome}

\usepackage[utf8]{inputenc}
\usepackage{booktabs}
\usepackage{graphicx}
\usepackage{array}
\usepackage{tabularx}
\usepackage{xcolor,colortbl}  
\definecolor{boxcolor}{HTML}{d92523} 
\definecolor{bulbcolor}{HTML}{e3b87f} 

\hypersetup{
    colorlinks=true,            
    linkcolor=MidnightBlue,     
    filecolor=magenta,          
    urlcolor=cyan,              
    citecolor=cvprblue,         
    pdftitle={Overleaf Example},
    pdfpagemode=FullScreen,
}
 
\renewcommand{\today}{}
\newcommandx{\info}[2][1=]{\todo[linecolor=red,backgroundcolor=red!25,bordercolor=red,#1]{#2}}

\title{\centering IV-Bench: A Benchmark for Image-Grounded Video\\ Perception and Reasoning in Multimodal LLMs}

\author{
\textbf{M-A-P} \\
ByteDance Inc.
}

\begin{abstract}
Existing evaluation frameworks for Multimodal Large Language Models (MLLMs) primarily focus on image reasoning or general video understanding tasks, largely overlooking the significant role of image context in video comprehension. To bridge this gap, we propose \textbf{IV-Bench}, the first comprehensive benchmark for evaluating \emph{Image-Grounded Video
Perception and Reasoning}. IV-Bench consists of 967 videos paired with 2,585 meticulously annotated image-text queries across 13 tasks (7 perception and 6 reasoning tasks) and 5 representative categories. Extensive evaluations of state-of-the-art open-source (e.g., InternVL2.5, Qwen2.5-VL) and closed-source (e.g., GPT-4o, Gemini2-Flash and Gemini2-Pro) MLLMs demonstrate that current models substantially underperform in image-grounded video Perception and Reasoning, merely achieving at most 28.9\% accuracy. Further analysis reveals key factors influencing model performance on IV-Bench, including inference pattern, frame number, and resolution. Additionally, through a simple data synthesis approach, we demonstratethe challenges of IV‑Bench extend beyond merely aligning the data format in the training proecss. These findings collectively provide valuable insights for future research. Our codes and data are released in \url{https://github.com/multimodal-art-projection/IV-Bench}.
\end{abstract}

\begin{document}
\begin{CJK*}{UTF8}{gbsn}

\maketitle

\begin{figure}[htbp]
    \centering
    \includegraphics[width=0.95\textwidth]{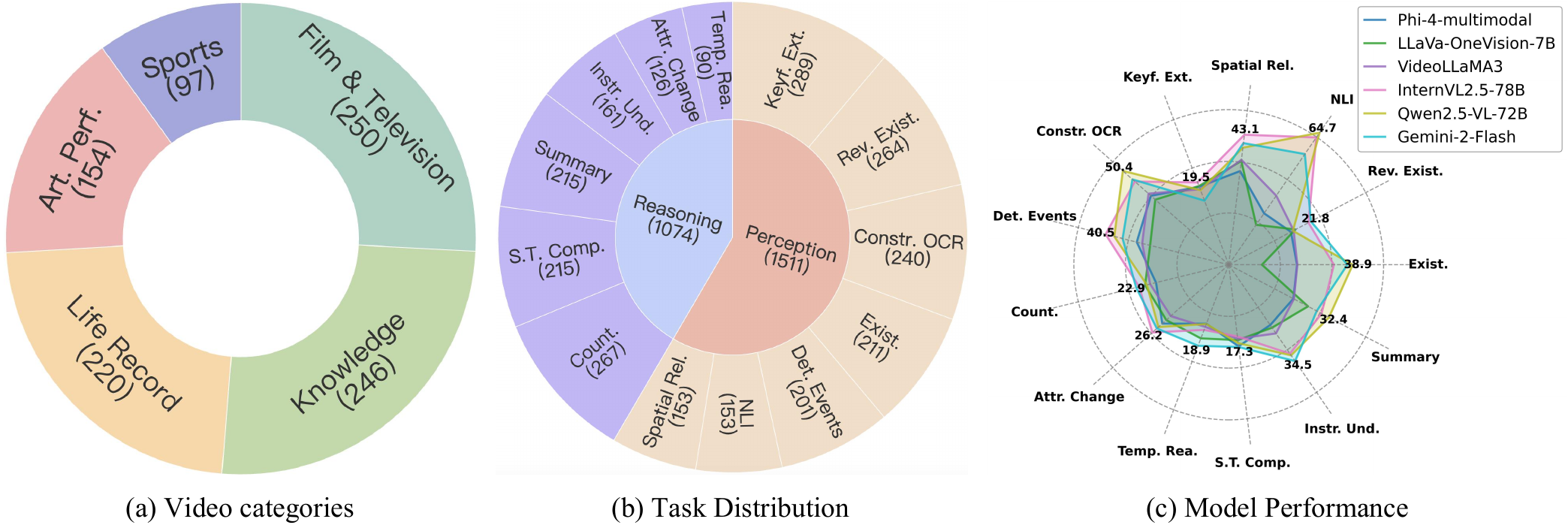}
    \caption{(a) Video Categories. IV-Bench includes videos spanning five representative categories, ensuring diverse topical coverage. (b) Task distribution in IV-Bench. IV-Bench consists of a total of 13 tasks, which are categorized into two main types: 6 reasoning tasks and 7 perception tasks. (c) Model Performance on IV-Bench. All evaluated MLLMs exhibit limited performance on IV-Bench. Even on the best-performing task (Natural Language Inference), the highest achieved accuracy is merely 64.7\%, with other tasks resulting in substantially lower scores.}

    \label{fig:overview}
\end{figure}

\newpage

\begingroup
\setlength{\parskip}{5.7pt}
\tableofcontents
\endgroup

\newpage

\section{Introduction}


Building upon the remarkable success of Large Language Models (LLMs) across various AI tasks \citep{young2024yi, zhang2024map}, Multimodal Large Language Models (MLLMs) have shown impressive capabilities in integrating and interpreting information from multiple modalities, such as text, images, and videos \citep{liu2023visual, chen2024internvl, guo2024mammoth}. Consequently, diverse benchmarks \citep{yue2024mmmu,zhang2024cmmmu,hu2025video,cheng2025simplevqa,wu2024scimmir,wu2024mmra,zhu2024limemllmevaluation} have emerged to systematically evaluate their multimodal integration and task-solving capabilities.


Images capture subtle details, such as facial features, that are difficult to accurately describe in text, especially when a person's identity is unknown. For instance, if we lack a name, text descriptions may resort to vague clues like clothing color or hair color, which might not be sufficient to uniquely identify someone. In contrast, images offer complete visual cues that clearly depict these details. However, existing video benchmarks typically focus on general video comprehension with purely text-based queries, neglecting the critical scenario where static images provide essential context for video understanding~\citep{fang2025mmbench,li2024mvbench}. Image-grounded video perception and reasoning\textemdash the ability to leverage a image as critical contextual information to locate and interpret video content\textemdash is fundamental for numerous real-world applications, such as accurate scene interpretation, object recognition, and event retrieval. Despite its importance, there is currently no benchmark specifically designed to evaluate this capability.

To address this critical research gap, we introduce \textbf{IV-Bench}, \textit{the first comprehensive benchmark for evaluating MLLMs in image-grounded video perception and reasoning tasks}. IV-Bench comprises 967 videos paired with 2,585 meticulously annotated image-text queries, spanning 13 distinct tasks (7 perception and 6 reasoning tasks) across five representative categories (Knowledge, Film \& Television, Sports Competition, Artistic Performance, and Life Record), as shown in Figure \ref{fig:overview}(a). Notably, the images used in IV-Bench are sourced externally, not extracted from the videos, ensuring the generalizability and robustness of the benchmark.

We conduct extensive evaluations on 23 state-of-the-art open-source models (e.g., InternVL2.5 \citep{chen2024expanding}, Qwen2.5-VL series \citep{qwen2.5-VL}) and 4 closed-source models (e.g., Doubao-1.5-vison-pro\citep{doubao1.5pro}, GPT-4o~\citep{gpt4o}, Gemini-2 Flash \citep{team2024gemini}, Gemini-2 Pro \citep{team2024gemini}). Results demonstrate that existing models substantially struggle with image-grounded video perception and reasoning tasks, with the top-performing MLLMs achieving only 28.9\% accuracy. Performance deteriorates further on complex reasoning tasks such as Temporal Reasoning, underscoring significant limitations and highlighting a critical research gap.

Furthermore, ablation studies comparing models with and without image contexts reveal that smaller models show minimal benefit from incorporating images, whereas larger models significantly benefit from image contexts, particularly when images follow the video. Additional analysis identify other key factors influencing model performance, providing valuable insights for future research.

To investigate whether the lack of video image formatted training data contributes to the performance gap on IV‑Bench, we employ a simple synthetic data pipeline that automatically generates supervised fine‑tuning examples from existing video QA datasets. Although this automated augmentation yields slight improvements, the gains remain minimal—suggesting that the challenges of IV-Bench arise from deeper image‑grounded video understanding ability demands rather than mere data format alignment. We hope these results will motivate the development of more advanced methods for tackling image‑grounded video perception and reasoning.

In summary, our work has three-fold contributions:
\begin{itemize}
    \item \textbf{IV-Bench}. We present IV-Bench, the first comprehensive benchmark for image-grounded video perception and reasoning in MLLMs. IV-Bench comprises 967 videos paired with 2,585 meticulously annotated image-text queries, where the images, collected from external sources rather than extracted from the videos themselves, provide the essential context required to accurately answer the queries. The dataset spans 5 major categories and covers 13 distinct tasks (7 perception and 6 reasoning tasks), ensuring substantial diversity across various scenarios and task types. Moreover, two round quality control—one ensuring clarity, accuracy, and category labeling, and another confirming that both image and video are required to answer correctly, ensuring the high quality of IV-Bench.
    \item \textbf{Comprehensive Evaluation of MLLMs}. We evaluate 27 state-of-the-art MLLMs, including the latest closed-source models (e.g., GPT-4o, Gemini-2-Flash and Gemini-2-Pro) and open-source models (e.g., InternVL2.5 and Qwen2.5-VL series). Our experiments reveal that current models perform sub-optimally on image-grounded video perception and reasoning, with the best model achieving only 28.9\% overall accuracy and just 24.9\% on reasoning tasks—clearly indicating an urgent need for enhanced image-grounded video perception and reasoning capabilities in MLLMs.
    \item \textbf{Insights for Future Research}. Our analysis provides key insights to guide future research. Ablation studies indicate that increasing frame number and video resolution positively affect performance. Moreover, larger models significantly benefit from image contexts presented after the video, while smaller models show minimal improvements. Through a simple data synthesis approach, we demonstrate that the challenges of IV-Bench do not arise from a lack of video–image format alignment in training data, which underscoring the need for more advanced methods beyond mere format alignment.
\end{itemize}

\section{IV-Bench}

\begin{figure*}[t]
    \centering
    \includegraphics[width=\textwidth]{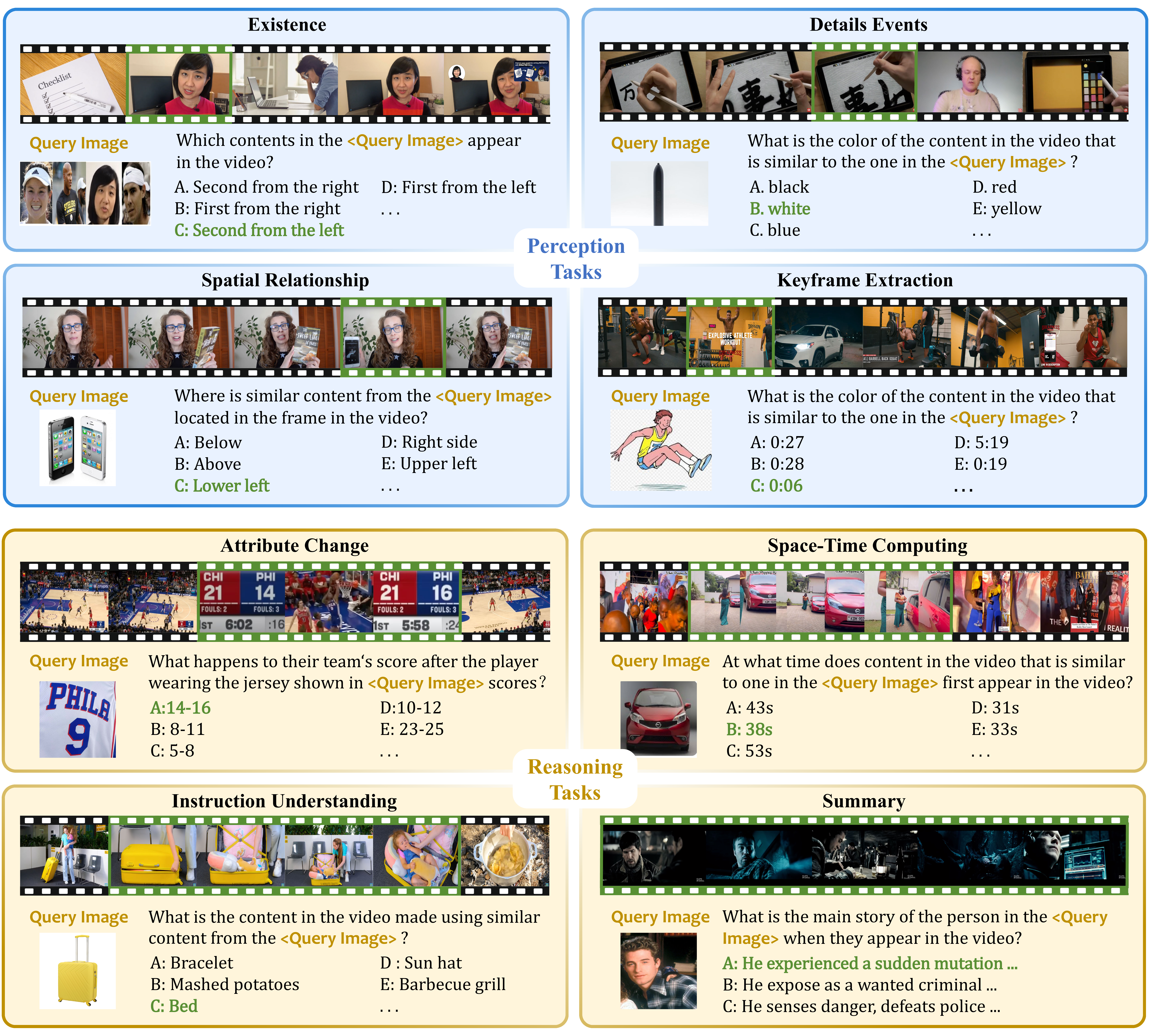}
    \caption{Representative examples from IV-Bench. Each sample consists of a video paired with an image-text query, comprising a query image and corresponding query text. The correct answer is marked in \textcolor[RGB]{89,143,52}{green}, with relevant video frames also highlighted in green.}
    \label{fig:tasks}
\end{figure*}

\subsection{Overview}
IV-Bench is designed to evaluate image-grounded video perception and reasoning, aiming to assess the capabilities of MLLMs in utilizing external visual cues for localization, reasoning, and comprehension of video content. IV-Bench consists of 967 diverse videos paired with 2,585 image-text queries, with each image providing indispensable contextual cues necessary for correctly answering the queries. The dataset spans 13 distinct tasks. Key features of IV-Bench include:

\begin{itemize}
    \item \textbf{Image-Text Queries}. For each video, we design multiple image‑text queries. Each query includes an externally sourced image—not extracted from the video itself—and an associated textual question. These externally sourced images guarantee greater visual diversity and better simulate real‑world usage scenarios, providing critical contextual cues necessary to accurately answer the queries.
    \item \textbf{Diverse Video Categories}. The dataset covers a wide array of categories including Knowledge, Film \& Television, Sports Competitions, Artistic Performances, and Life Records, ensuring extensive content diversity for various research purposes. Each video, with a minimum duration of five minutes, provides sufficient depth for comprehensive analysis.
    \item \textbf{Diverse Evaluation Tasks}. As illustrated in Figure \ref{fig:overview}(b), IV-Bench offers 13 distinct evaluation tasks grouped into perception and reasoning categories. These tasks comprehensively assess the capability of MLLMs to perform image-grounded video perception and reasoning, spanning a diverse set of perceptual and reasoning skills.
\end{itemize}

\subsection{Task Definitions}

IV-Bench comprises 7 perception tasks and 6 reasoning tasks, with representative examples of selected tasks shown in Figure \ref{fig:tasks}. Additional examples of the remaining tasks are provided in Figure~\ref{fig:remaining_tasks}.
These tasks address various aspects of image-grounded video comprehension, spanning from basic perception to complex reasoning. Detailed descriptions of each task are presented below.

\noindent\textbf{Perception Tasks:} These tasks evaluate the model's capability to directly extract and interpret fundamental visual information from the video. They primarily focus on recognizing objects, people, scenes, and spatial relationships by leveraging contextual cues from the reference image. In essence, perception tasks assess the model's ability to accurately "see" and identify content within the video.

\begin{itemize}
    \item Existence: Identify which objects or people in the reference image appear in the video.
    \item Reverse Existence:  Identify objects or people present in the image but absent in the video.
    \item Natural Language Inference (NLI): Determine which scenes in the video are similar to a specific scene in the image.
    \item Spatial Relationship: Identify absolute or relative spatial relationships among objects or people in the video, grounded by the reference image.
    \item Keyframe Extraction: Identify precise timestamps or segments within the video where objects or people depicted in the image appear.
    \item Constrained OCR: Recognize text-based content in the video, constrained by conditions explicitly defined by the reference image, such as spatial alignment, temporal correspondence, and semantic relevance.
    \item Detailed Events: Identify specific events or actions within the video directly related to content depicted in the reference image.
\end{itemize}

\noindent\textbf{Reasoning Tasks:} These tasks require models to engage in higher-order cognitive functions by integrating visual cues with contextual and temporal information. They assess the model's capacity to analyze, synthesize, and infer meaningful conclusions beyond simple visual recognition.

\begin{itemize}
    \item Counting: Count occurrences of a person, object, or action depicted in the video, grounded by the reference image.
    \item Space-Time Computing: Calculate event durations or distances between objects/people in the video, using the image as contextual guidance.
    \item Summary: Generate a brief description summarizing a person, object, or event depicted in the video, informed by the reference image.
    \item Instruction Understanding: Understand the functionality or creation process of objects depicted in the video, guided by the reference image.
    \item Attribute Change: Detect changes in attributes (e.g., clothing, size, color) of objects or people throughout the video, referenced by the image.
    \item Temporal Reasoning: Infer precise start and end timestamps of target events using temporal cues and world knowledge, as introduced in~\citep{huang2024lita}.
\end{itemize}

\subsection{Annotation and Quality Control}

The high quality of IV-Bencg stems from a rigorous annotation protocol and two‑stage quality control: annotators review each video, assign one task type, then select or create external images exclusively from non‑video sources and formulate questions with up to nine plausible distractors. In the first round quality control , we verify question clarity, answer accuracy, and label consistency; and in the second round, we remove any items solvable by video or common sense, eliminate visual‑information leakage, and ensure that each sample includes at least two "effective" distractors that, although incorrect for the current image, would become the correct answer to a different question with the identical text query but a different image, thereby ensuring each sample’s image necessity.

\subsubsection{Annotation Process}

\begin{itemize}
    \item \textbf{Video Collection}: A total of 976 videos, each exceeding five minutes in length, are carefully selected to ensure broad topical coverage. These videos span five distinct categories: Knowledge, Film \& Television, Sports, Artistic Performances, and Life Records, providing diverse content suitable for various research applications.
    
    \item \textbf{Task Assignment}: Annotators first watch each video in its entirety before performing image and text annotations. They then assign an appropriate task type from 13 predefined categories (detailed below). This initial task assignment guides subsequent annotations, ensuring alignment with task-specific requirements.
    
    \item \textbf{Image and Question Annotation}: After task assignment, annotators manually retrieve a relevant external image from online sources, explicitly ensuring it is not extracted from the video itself. The selected image must be closely related to specific keywords, individuals, or themes present in the video. Annotators then formulate a text question leveraging both the video content and contextual cues provided by the image.
    
    \item \textbf{Answer and Distractor Design}: Annotators craft the correct answer by carefully analyzing multimodal information from both the video and the image. Additionally, they generate up to nine plausible yet incorrect distractors to increase the question's difficulty while ensuring contextual relevance. For certain questions, fewer distractors may be provided depending on content constraints.

\end{itemize}

\subsubsection{Quality Control}
Ensuring the quality and consistency of the data is crucial for creating a reliable dataset. Our quality control process is conducted in two main rounds (see appendix \ref{fig:tasks_qc} for more detail of quality control)：

\begin{itemize}
    \item \textbf{First Round Quality Control}: The First round quality check focuses on the structure and content standardization of evaluation questions. It mainly checks whether the query and options are clearly described, whether the answer is correct, and whether the distractors can effectively mislead test-takers. So we verify the clarity, precision, and unambiguity of each question. We also confirm that the correct answers and distractors are both plausible and contextually relevant, ensuring each query can be accurately answered based on the provided video and image. Furthermore, we check task categorization accuracy, correcting any misclassifications to maintain consistency across all 13 predefined tasks.
    
    \item \textbf{Second Round Quality Control}: Since some questions can be answered using only common sense or video content—we conduct a second round of quality check. During this phase, any query that can be resolved without the reference image or video is simply removed, and any text query that inadvertently reveals visual content is rewritten to eliminate leakage. We also pinpoint ineffective distractors—those easily dismissed using video alone—and manually introduce at least two effective distractors per question; These distractors are crafted so that, although incorrect for the current image, they would serve as the correct answers to alternative questions sharing the same text query but paired with a different image—thereby ensuring that the image is necessary for each sample in IV-Bench.
\end{itemize}

Two rounds of rigorous quality control—verifying question clarity, answer correctness, prevention of image‐information leakage, and the inclusion of effective distractors—substantially bolster the integrity of IV-Bench. Collectively, these measures establish IV‑Bench as a high‑quality dataset for advancing image‑grounded video perception and reasoning research.



\section{ Comparison with other video benchmarks}
\begin{table*}[t]
\centering
\caption{A comparison of representative video benchmarks. Benchmarks are categorized based on query modality into text-only benchmarks and image-text benchmarks. IV-Bench is the first manually annotated benchmark explicitly designed to evaluate image-grounded video perception and reasoning, comprising 7 perception tasks and 6 reasoning tasks. ImgSrc, ImgNec, and VidNec are abbreviations for Image Source, Image Necessity, and Video Necessity, respectively.}
\resizebox{\textwidth}{!}{%
\begin{tabular}{ccccccccccc}
\toprule
\textbf{Query Modality} & \textbf{Benchmark} & \textbf{\#Videos} & \textbf{Duration. (s)} & \textbf{\#Tasks} & \textbf{\#QA Pairs} & \textbf{Anno.} & \textbf{\# Avg. Opt.}  & \textbf{ImgSrc} & \textbf{ImgNec} & \textbf{VidNec} \\ \midrule
\multirow{9}{*}{Text} & MMBench-Video & 609 & 165 & 26 & 1,998 & M & - & - & - & \textcolor{green}{\ding{51}} \\
        & Video-Bench & 5,917 &  56 & 10 & 17,036  &  M \& A &  4 & - & - & \textcolor{green}{\ding{51}} \\
        & EgoSchema &  5,063 & 180 & - & 5,063 & M \& A & 5 & - & - & \textcolor{green}{\ding{51}} \\
        & AutoEval-Video & 327 & 14.6 & - & 327 & A & - & - & - & \textcolor{green}{\ding{51}} \\
        & TempCompass & 410 & 11.4 & - & 7,540 & M \& A & - & - & - & \textcolor{green}{\ding{51}} \\
        & MVBench & 3,641 & 16 & 20 & 4,000 & M & 4 & - & - & \textcolor{green}{\ding{51}} \\ 
        & LongVideoBench & 3,763 & 473 & 17 & 6,678 & M & 4 & - & - & \textcolor{green}{\ding{51}} \\ 
        & MLVU & 1,730 & 930 & 9 & 3,102 & M \& A & 4 or 6 & - & - & \textcolor{green}{\ding{51}} \\ 
        & Video-MME & 900 & 1,017.9 & 12 & 2,700 & M & 4 & - & - & \textcolor{green}{\ding{51}} \\ \hline
         
Text+Image    & V2P-Bench    & 980 & 1140   & 12 & 1172 & M &  4 & In-Video & \textcolor{green}{\ding{51}} & \textcolor{green}{\ding{51}} \\
Text-only/Text+Image & Video‑MMMU & 300 & 506.2 & 6  & 900   & M & 10 & Out-of-Video & \textcolor{red}{\ding{55}} & \textcolor{red}{\ding{55}} \\
Text+Image    & IV‑Bench    & 967 & 537   & 13 & 2,585 & M &  9 & Out-of-Video & \textcolor{green}{\ding{51}} & \textcolor{green}{\ding{51}} \\
\bottomrule
\end{tabular}}

\end{table*}
As shown in Table 1, existing approaches can be broadly categorized into two groups: benchmarks with text-only queries and those with combined image-text queries.

Benchmarks with text-only queries primarily emphasize evaluating video understanding guided solely by textual instructions. For instance, benchmarks such as MMBench-Video \citep{fang2025mmbench}, Video-Bench \citep{ning2023video}, and four others mainly target short-video analysis tasks (typically under 180 seconds), exhibiting notable limitations in evaluating long-range temporal reasoning. Benchmarks such as LongVideoBench \citep{wu2025longvideobench}, MLVU \citep{zhou2024mlvu}, and Video-MME \citep{hu2025video} broaden evaluation scope by constructing corpora comprising videos of varied lengths, with Video-MME specifically averaging around 500 seconds. Notably, Video-MME innovatively incorporates additional evaluation dimensions, including subtitle recognition and audio understanding. However, none of these benchmarks employ image-text queries, thus limiting their capability to evaluate image-grounded video perception and reasoning.

Comprising 980 videos and 1,172 visual‑prompt QA pairs, V2P‑Bench \citep{zhao2025v2p} encompasses five principal tasks and twelve specialized dimensions for the assessment of fine‑grained video comprehension via visual prompts. Nonetheless, deriving visual prompts solely from video frames limits both the diversity of visual content and the range of real‑world usage scenarios. Among video benchmarks employing image–text queries, Video‑MMMU \citep{hu2025video} is the only one to leverage external images—i.e. images sourced from out of video—to assess the ability of MLLMs to acquire knowledge from videos. In contrast, IV-Bench is the first manually annotated benchmark explicitly created for image-grounded video perception and reasoning, with each query meticulously formulated to ensure the image information is essential for deriving the correct answer. IV‑Bench differs from Video‑MMMU by making the video indispensable for every query, pairing text with an image in every query, and using effective distractors to ensure that the image is always necessary for each query. Refer to the Appendix for illustrative examples that compare IV-Bench and VideoMMMU.

\section{Experiments}
In this section, we evaluate multiple representative MLLMs using IV-Bench. We first introduce the evaluated models and experimental settings, followed by a quantitative comparison of performance between open-source and commercial (closed-source) models. We then analyze the impact of various factors on model performance, including frame number, resolution, and inference patterns. Finally, we propose a simple synthetic data approach to further validate the inherent difficulty and quality of IV-Bench.

\subsection{Settings}
We evaluate 4 commercial models: Doubao-1.5-vision-pro~\citep{doubao1.5pro}, GPT-4o~\citep{gpt4o} Gemini 2 Flash~\citep{team2024gemini}, and Gemini 2 Pro~\citep{team2024gemini}. For open-source models, we select 23 representative MLLMs, including the Qwen2.5-VL series, InternVL2.5-VL series, and VideoLLaMA3~\citep{zhang2025videollama}. We employ a uniform sampling strategy to process video frames. For all models, we uniformly set the frame number to 32. The default evaluation input format is "video frames + image + question" with prompts, indicating that video frames are provided first, followed by the image. Since test samples in IV-Bench are multiple-choice questions with 10 options, we adopt accuracy as the evaluation metric, where random guessing accuracy is 10\%. Accuracy is computed by directly matching the model’s output to the correct answer.


\subsection{Main Results}
\begin{table*}[t]
  \centering
  \caption{The performance of MLLMs on IV-Bench across 13 tasks, comprising 7 perception tasks and 6 reasoning tasks. For each task, the best performance is indicated in bold, and the second-best performance is indicated with underlining. Note that "\textbf{P-Avg}" and "\textbf{R-Avg}" denote the average results on perception and reasoning tasks, respectively.}
  \label{tab:llm_comparison}
  \resizebox{\textwidth}{!}{%
  \begin{tabular}{l|c|cccccccc|ccccccc|c}
    \hline
    \toprule
    \multirow{2}{*}{\textbf{Models}} & \multirow{2}{*}{\textbf{Overall}} & \multicolumn{8}{c|}{\textbf{Perception}} & \multicolumn{7}{c}{\textbf{Reasoning}} \\
    \cmidrule(lr){3-10} \cmidrule(lr){11-17}
    & & \textbf{Exist.} & \textbf{RE} & \textbf{NLI} & \textbf{SR} & \textbf{KE} & \textbf{CO} & \textbf{DE} & \textbf{P-Avg} & \textbf{Cnt.} & \textbf{AC} & \textbf{TR} & \textbf{STC} & \textbf{IU} & \textbf{Sum.} & \textbf{R-Avg} \\
    \midrule
    \multicolumn{17}{l}{\textit{Open Source MLLMs($<$ 10B)}} \\
    \midrule
    Llama-vid~\citep{li2024llama} & 10.5 & 13.3 & 10.6 & 6.5 & 7.2 & 9.8 & 13.3 & 16.0 & 11.2 & 7.1 & 8.7 & 2.2 & 12.2 & 11.2 & 12.3 & 9.5 \\
    LLaVA-Mini~\citep{zhang2025llava}  & 12.5 & 7.1 & 12.1 & 9.8 & 15.0 & 10.8 & 12.9 & 18.0 & 12.1 & 12.7 & 13.5 & 13.3 & 14.0 & 14.7 & 11.3 & 13.1 \\
    MAmmoTH-VL-8B~\citep{guo2024mammoth} & 13.3 & 3.3 & 8.7 & 3.9 & 29.4 & 10.5 & 19.6 & 14.5 & 12.4 & 12.7 & 16.7 & 7.8 & 13.1 & 14.0 & 20.3 & 14.5 \\
    Longva~\citep{zhang2024long}  & 14.4 & 5.2 & 20.1 & 4.6 & 20.3 & 13.9 & 27.5 & 19.5 & 16.4 & 7.5 & 11.9 & 14.4 & 15.0 & 10.5 & 13.2 & 11.7 \\
    NVILA~\citep{liu2024nvila}  & 14.4 & 2.4 & 17.4 & 2.6 & 22.2 & 13.2 & 24.6 & 14.0 & 14.2 & 12.7 & 18.3 & 18.9 & 17.3 & 14.7 & 10.8 & 14.7 \\
    Longvu~\citep{shen2024longvu}  & 14.8 & 3.8 & 17.1 & 11.1 & 20.9 & 11.5 & 19.6 & 19.5 & 14.7 & 18.0 & 16.7 & 8.9 & 15.9 & 16.8 & 10.9 & 15.0 \\
    Phi-3.5-vision~\citep{phi-3.5-mini-instruct}  & 15.2 & 5.2 & 12.9 & 11.1 & 23.5 & 10.8 & 25.0 & 22.5 & 15.5 & 16.1 & 15.9 & 11.1 & 13.6 & 14.7 & 15.6 & 14.8 \\
    Phi-4-multimodal~\citep{abdin2024phi}  & 16.6 & 11.8 & 12.5 & 9.8 & 22.4 & 17.5 & 27.2 & 22.6  & 17.8 & 14.1 & 19.8 & 10.0 & 17.0 & 13.5 & 13.5 & 14.8 \\
    LLaVA-OneVision-7B~\citep{li2024llavaonevisioneasyvisualtask} & 16.3 & 2.8 & 14.4 & 5.9 & 27.5 & 17.8 & 24.2 & 16.5 & 15.7 & 18.7 & 17.5 & 15.6 & 14.5 & 14.7 & 20.3 & 17.2 \\
    InternVL2-8B~\citep{OpenGVLab_InternVL2-8B}  & 16.8 & 7.1 & 10.6 & 19.0 & 25.5 & 12.2 & 27.9 & 22.5 & 17.1 & 18.7 & 20.6 & 6.7 & 14.5 & 14.7 & 17.9 & 16.3 \\
    VAMBA~\citep{ren2025vamba} & 16.9 & 13.7 & 11.7 & 13.7 & 26.1 & 12.9 & 27.9 & 22.0 & 17.8 & 15.7 & 21.4 & 12.2 & 13.6 & 13.3 & 16.5 & 15.5 \\
    Minicpm-v~\citep{yao2024minicpm} & 17.2 & 7.1 & 15.5 & 9.2 & 30.1 & 14.3 & 32.5 & 22.5 & 18.6 & 15.7 & 15.9 & 12.2 & 15.0 & 13.3 & 17.5 & 15.3 \\
    Minicpm-o~\citep{yao2024minicpm} & 17.1 & 8.1 & 14.8 & 8.5 & 28.8 & 14.6 & 31.2 & 23.5 & 18.4 & 13.1 & 13.5 & 12.2 & 17.8 & 16.8 & 16.5 & 15.2 \\
    VideoLLaMA3~\citep{zhang2025videollama} & 17.4 & 11.9 & 13.6 & 17.7 & 28.1 & 16.4 & 28.9 & 20.0 & 19.0 & 16.5 & 15.1 & 11.1 & 14.0 & 17.5 & 13.7 & 14.9 \\
    InternVL2.5-8B~\citep{chen2024expanding} & 17.4 & 5.4 & 15.7 & 16.4 & 28.3 & 10.1 & 31.4 & 25.2 & 17.8 & 15.7 & 22.1 & 8.2 & 14.1 & 16.3 & 20.6 & 16.5 \\
    Qwen2.5-VL-7B~\citep{qwen2.5-VL} & 18.5 & 8.1 & 13.6 & 21.6 & 26.8 & 15.7 & 31.7 & 18.5 & 18.9 & 16.5 & 23.0 & 6.7 & 15.4 & 17.5 & 24.1 & 17.9 \\
    \midrule
    \multicolumn{17}{l}{\textit{Open Source MLLMs($>$ 10B)}} \\
    \midrule
    Aria~\citep{li2024aria} & 17.4 & 9.5 & 9.5 & 11.8 & 24.8 & 17.4 & 36.7 & 20.5 & 18.6 & 16.9 & 16.7 & 15.6 & 14.0 & 18.9 & 13.7 & 15.8 \\
    InternVL2.5-26B & 20.6 & 12.8 & 14.4 & 33.3 & 36.6 & 13.6 & 32.5 & 24.5 & 22.4 & 15.0 & 23.0 & 14.4 & 14.0 & 25.2 & 19.8 & 18.1 \\
    LLaVA-OneVision-72B & 22.3 & 15.6 & 18.9 & 40.5 & 28.8 & \textbf{20.1} & 32.9 & 27.0 & 25.2 & 17.6 & 17.5 & \underline{18.9} & 15.3 & 21.1 & 20.5 & 18.3 \\
    Qwen2.5-VL-32B & 23.0 & 13.3 & \underline{22.8} & 28.8 & 31.6 & 13.8 & 41.7 & 27.1 & 24.9 & 20.5 & 23.0 & 12.2 & 14.7 & 25.4 & 23.6 & 20.2 \\
    InternVL2.5-38B & 26.6 & 24.6 & 19.3 & 56.9 & 34.0 & 19.5 & 38.8 & 33.5 & 30.4 & 20.2 & 22.2 & 6.7 & \textbf{22.0} & 27.3 & 22.6 & 21.1 \\
    InternVL2.5-78B & \underline{28.6} & 28.0 & 20.5 & \underline{60.1} & \textbf{43.1} & \underline{19.5} & 39.6 & \textbf{40.5} & \underline{33.4} & \underline{21.4} & \textbf{26.2} & 12.2 & 13.1 & 29.4 & \underline{27.8} & 21.9 \\
    Qwen2.5-VL-72B & \textbf{28.9} & \textbf{38.9} & 13.3 & \textbf{64.7} & 34.9 & 16.0 & \textbf{50.4} & \underline{35.0} & \textbf{33.7} & 17.9 & 22.2 & 10.1 & 15.7 & 30.5 & \textbf{32.4} & 21.9 \\
    \midrule
    \multicolumn{17}{l}{\textit{Closed Source MLLMs}} \\
    \midrule
    Doubao-1.5-vison-pro\citep{doubao1.5pro} & 19.2 & 20.9 & 12.5 & 19.0 & 26.1 & 14.7 & 31.7 & 26.5 & 21.0 & 12.8 & 23.0 & 17.8 & 14.4 & 15.1 & 20.1 & 16.5 \\
    GPT-4o~\citep{gpt4o} & 20.7 & 25.0 & 13.3 & 36.9 & 33.3 & 11.3 & 31.1 & 26.0 & 23.5 & 18.2 & 15.2 & 12.5 & 13.7 & 27.0 & 12.6 & 16.7 \\
    Gemini-2-Flash~\citep{team2024gemini} & 27.4 & \textbf{35.6} & 21.8 & 45.4 & 37.7 & 11.8 & 41.7 & 30.1 & 30.2 & \textbf{22.9} & \underline{23.2} & \textbf{18.9} & 17.3 & \underline{34.5} & 25.0 & \underline{23.4} \\
    Gemini-2-Pro~\citep{team2024gemini} & 27.7 & 26.4 & \textbf{23.8} & 35.5 & \underline{39.1} & 16.4 & \underline{43.8} & 30.1 & 29.6 & 22.6 & 21.6 & 15.6 & \underline{19.7} & \textbf{39.6} & 29.5 & \textbf{24.9} \\
    
    \bottomrule
  \end{tabular}%
}
\end{table*}

The performance across the 13 IV-Bench tasks—7 perception tasks and 6 reasoning tasks—is presented in Table \ref{tab:llm_comparison}. We report individual task performances along with average results for perception tasks (P-Avg) and reasoning tasks (R-Avg). From the results, we derive two primary conclusions:

\begin{itemize}
\item \textbf{Image-Grounded video perception and reasoning is Challenging}: Effectively performing image-grounded video perception and reasoning continues to pose significant challenges for MLLMs. For instance, among models under 10B parameters, the top‑performing Qwen2.5‑VL‑7B achieves merely 18.5\% accuracy, whereas random guessing would yield 11.11\%. This represents only a 7.39‑percentage‑point improvement over chance. Even larger models such as InternVL2.5-78B and Qwen2.5-VL-72B achieve just 28.6\% and 28.9\%, respectively. Moreover, the best-performing commercial model, Gemini-Pro, reaches only 27.7\%, highlighting significant untapped potential in leveraging image contexts to improve video comprehension. Overall, perception tasks generally prove easier than reasoning tasks. For example, InternVL2.5-78B achieves a perception task average (P-Avg) of 33.4\%, compared to a reasoning task average (R-Avg) of only 21.9\%. Even Gemini-Pro, which performs relatively better on reasoning tasks, manages only 24.9\%. Temporal reasoning remains especially challenging, with the top-performing model achieving only 16.7\% accuracy. Conversely, NLI and existence tasks appear relatively easier, with Qwen2.5-VL-72B scoring 38.9\% and 64.7\%, respectively.
\item \textbf{Moderate Gains with Larger Models}: Increasing model scale results in modest performance improvements across various tasks. For example, in the InternVL2.5 series, the 8B model achieves 17.4\% accuracy, while the larger 26B, 38B, and 78B variants reach 20.6\%, 26.6\%, and 28.6\%, respectively. Similarly, for the Qwen2.5-VL series, scaling from the 7B model (18.5\%) to the 72B model (28.9\%) results in noticeable performance gains. This improvement is particularly pronounced in few perception-based tasks. For example, Qwen2.5-VL-72B demonstrates a substantial 30.8\% improvement over Qwen2.5-VL-7B on existence tasks and a notable 43.1\% gain on NLI tasks. However, the benefits of scaling are notably smaller for reasoning-intensive tasks. For example, on counting tasks, Qwen2.5-VL-72B surpasses Qwen2.5-VL-7B by merely 1.4\%, while InternVL2.5-78B exhibits just a 5.7\% improvement over InternVL2.5-8B. The limited benefits observed on reasoning-intensive tasks may be attributed to the fact that increasing model size tends to enhance memorization and shallow pattern recognition more significantly than improves reasoning ability.
\end{itemize}


The experimental results confirm that image-grounded video perception and reasoning continue to pose significant challenges for MLLMs, especially for reasoning-intensive tasks. Furthermore, increasing model size yields only moderate improvements, suggesting that merely scaling models is insufficient to fully overcome these challenges. Future research should prioritize developing specialized mechanisms for video reasoning, such as enhanced temporal modeling techniques.

\subsection{Ablation Study}



\begin{figure*}[htbp]
    \centering
    \begin{subfigure}[b]{0.48\textwidth}
        \centering
        \includegraphics[width=\textwidth]{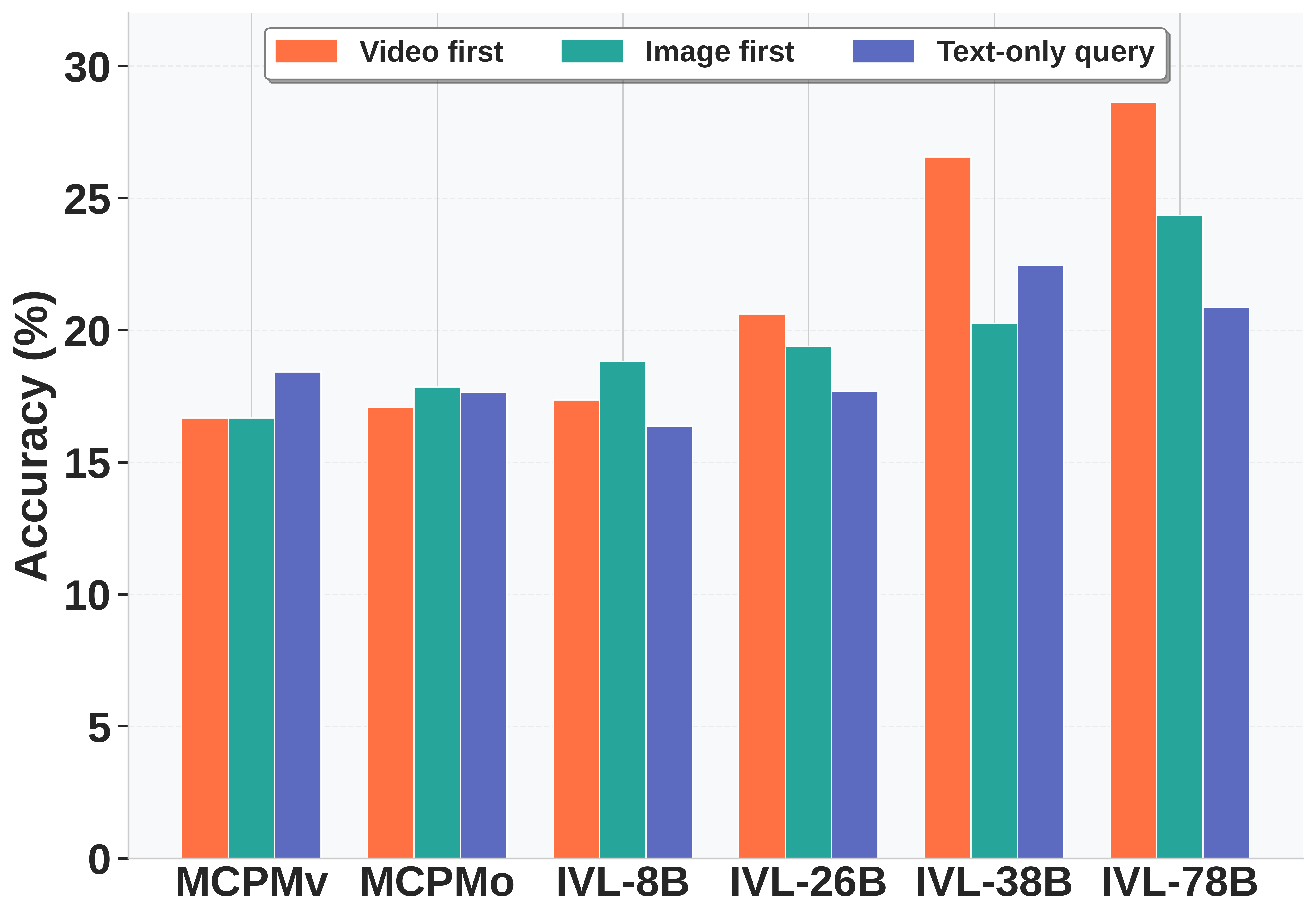}
        \captionsetup{justification=centering} 
        \caption*{(a)} 
        \label{fig:inference_pattern}
    \end{subfigure}
    \hfill
    \begin{subfigure}[b]{0.48\textwidth}
        \centering
        \includegraphics[width=\textwidth]{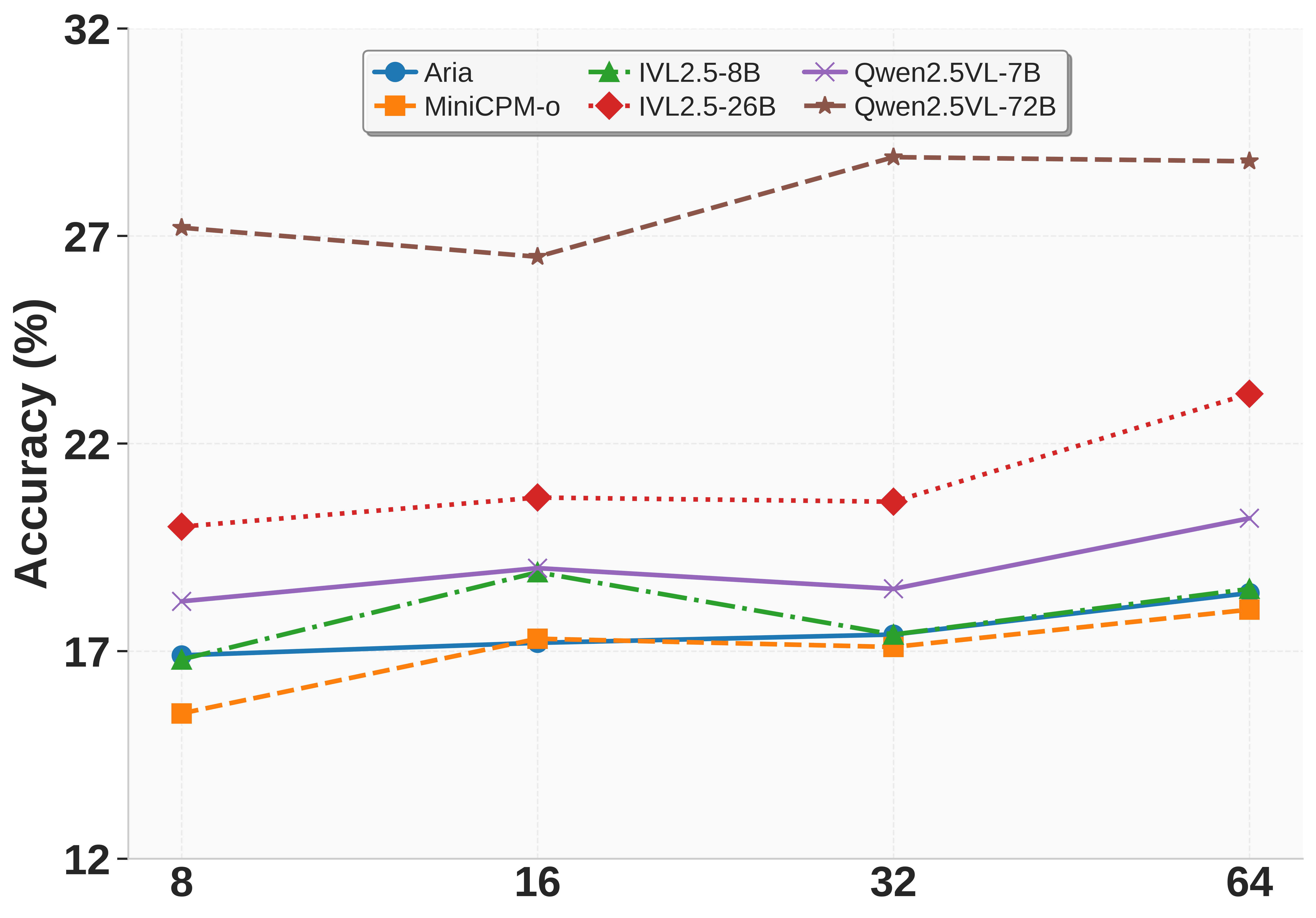}
        \captionsetup{justification=centering} 
        \caption*{(b)} 
        \label{fig:frame}
    \end{subfigure}
    \caption{Comparison of model performance: (a) across different inference patterns and (b) with varying numbers of frames. MCPMv/o represent MiniCPMv/o, IVL is the abbreviation of InternVL.}
    \label{fig:comparison}
\end{figure*}

We conduct in-depth analysis to investigate factors influencing model performance, including inference patterns, frame numbers, and video resolution.

\subsubsection{The Impact of Inference Pattern}

In this section, we comparatively analyze performance across three inference patterns: image-first, video-first (both under image-text query settings), and text-only queries (without image input). We evaluate both the MiniCPM-v/o and InternVL2.5 model series, with results presented in Figure ~\ref{fig:comparison}(a).

Our findings indicate that MiniCPM-v/o performs suboptimally under both image-first and video-first settings compared to the text-only query setting. One possible explanation is that insufficient image-grounded video perception and reasoning training data combined with limited generalization capacity means that adding image fails to enhance—and may even hinder—the ability to comprehend videos compared to text-only queries.
Similarly, InternVL2.5-8B demonstrates only minimal performance gains when incorporating images. Taken together, these observations suggest that \textbf{smaller models are less proficient in image grounded video perception and reasoning.}

Further analysis of the InternVL2.5‑VL models reveals that \textbf{larger models attain higher performance.} This implies a positive correlation between parameter size and capability for image-grounded video perception and reasoning. 
Additionally, for models such as InternVL2.5-26B, 38B, and 78B that effectively utilize image information, \textbf{placing the image after video frames results in superior performance compared to positioning it beforehand.} This phenomenon likely arises because images positioned at the beginning tend to be "forgotten" or overlooked by the model, causing the neglect of critical visual information that affects overall performance. A similar observation is presented in~\citep{ma2024vista}.

\subsubsection{Analysis of the Number of Visual Tokens}

This section addresses two key questions regarding the number of visual tokens supplied to multimodal vision–language models:

\begin{itemize}
  \item \textbf{Scaling Effect.} How does model performance change as we increase the total number of visual tokens—by varying either frame number or resolution?
  \item \textbf{Token Allocation.} When the total number of visual tokens is held constant, does allocating tokens to more frames or to higher resolution produce greater gains?
\end{itemize}

To answer these questions, we first conduct ablations isolating frame number and resolution, then evaluate seven frame–resolution pairs that yield approximately the same number of visual tokens to disentangle temporal versus spatial contributions.

\paragraph{Isolated Scaling of Frame number and Resolution}

In this subsection, we separately examine the effects of increasing frame number and resolution on model performance. We select six top-performing models—Aria, MiniCPM‑O, InternVL-8B/26B, and Qwen2.5‑VL-7B/72B. For the temporal study, we vary the number of frames (8, 16, 32, 64) while holding resolution fixed; the results are shown in Figure~\ref{fig:comparison}(a). We observe that \textbf{model performance consistently improves as frame number increases}, demonstrating that allocating additional visual tokens over time effectively enhances image‑grounded video perception and reasoning.

\begin{figure*}[htbp]
    \centering
    \begin{subfigure}[b]{0.48\textwidth}
        \centering
        \includegraphics[width=\textwidth]{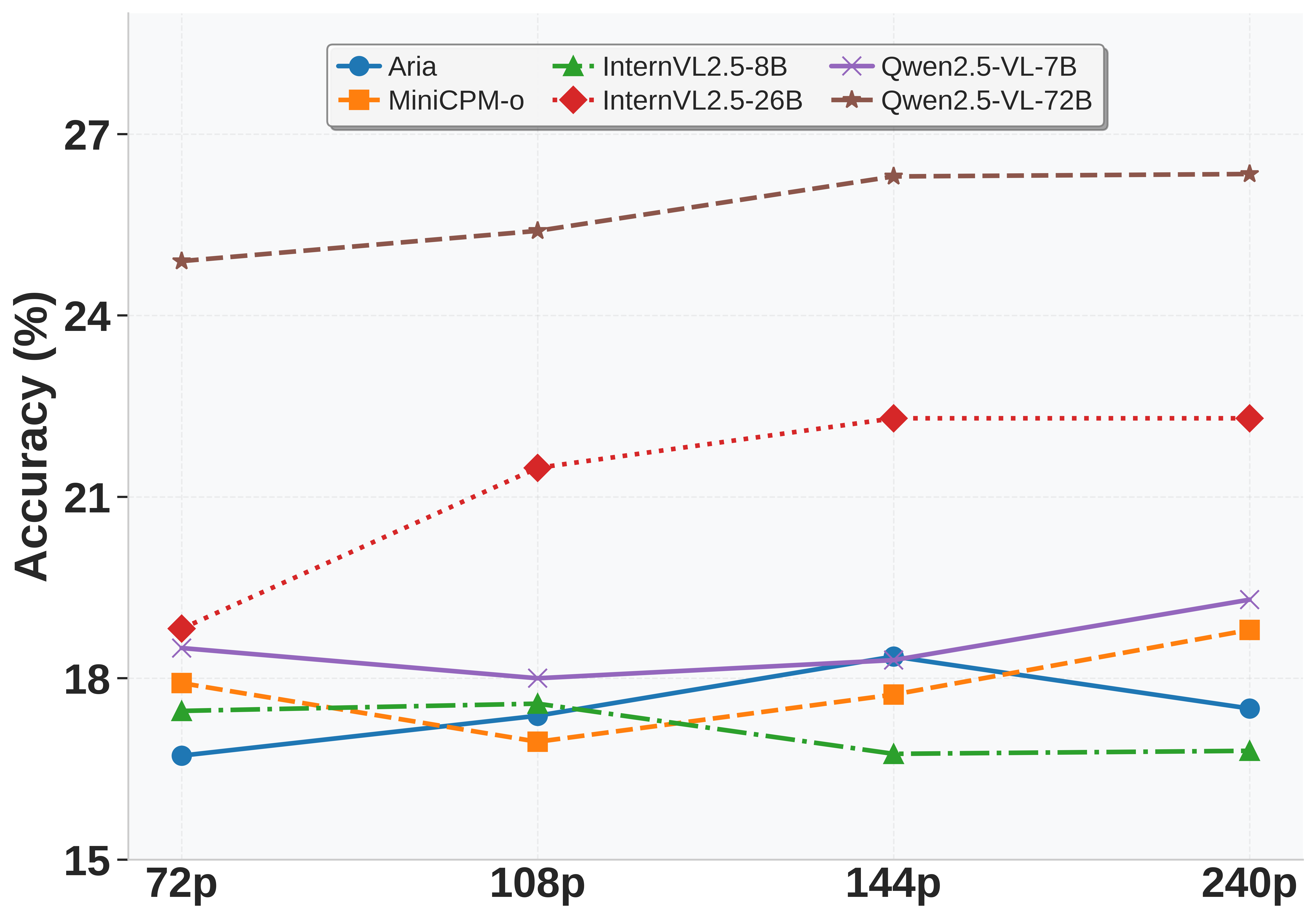}
        \captionsetup{justification=centering} 
        \caption*{(a)} 
        \label{fig:resolution}
    \end{subfigure}
    \hfill
    \begin{subfigure}[b]{0.48\textwidth}
        \centering
        \includegraphics[width=\textwidth]{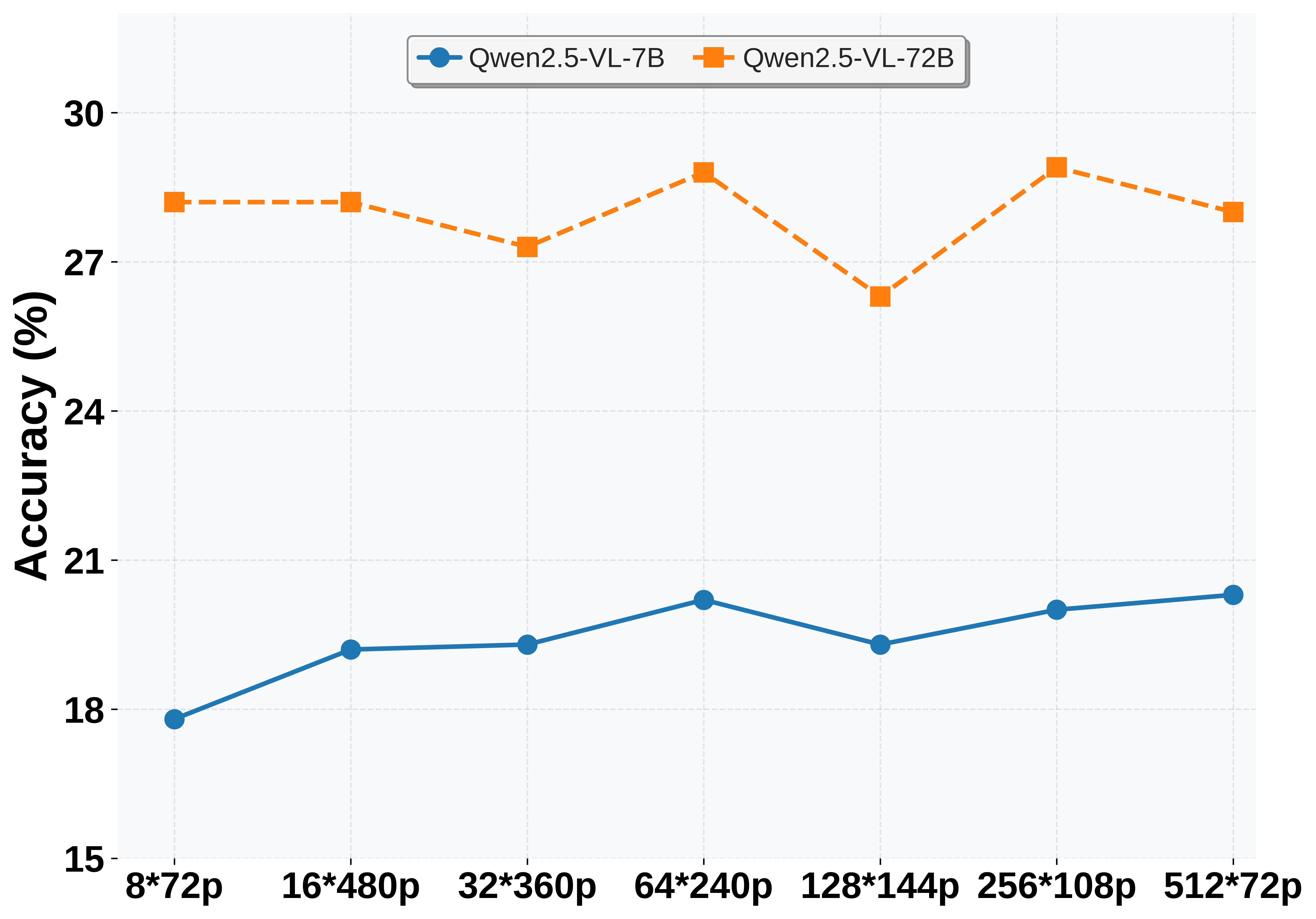}
        \captionsetup{justification=centering} 
        \caption*{(b)} 
        \label{fig:trade_off}
    \end{subfigure}
    \caption{Comparison of model performance: (a) across different video resolutions and (b) across various frame-resolution combinations.}
    \label{fig:comparison2}
\end{figure*}

For the spatial study, we fix the frame number at 32 and evaluate four resolutions (72p, 108p, 144p, 240p). As illustrated in Figure \ref{fig:comparison2}(a), \textbf{performance improves consistently as resolution increases—most markedly in low resolutions}—underscoring the benefit of allocating additional visual tokens to spatial detail for enhancing image‑grounded video perception and reasoning. \textbf{This rate of improvement diminishes at higher resolutions;} for example, upgrading from 144p to 240p yields gains in only two of the six models, indicating that, at 32 frames, further spatial token allocation beyond a mid‑range resolution offers only marginal benefit.

\paragraph{Token Allocation: Frames vs.\ Resolution}

Finally, to disentangle temporal and spatial contributions under a fixed budget of visual tokens, we evaluate seven frame–resolution pairs—(8, 720p), (16, 480p), (32, 360p), (64, 240p), (128, 144p), (256, 108p), and (512, 72p)—chosen to yield roughly equivalent token number. Figure~\ref{fig:comparison2}(b) reveals contrasting behaviors between model scales. For Qwen2.5‑VL‑7B, performance rises primarily with frame number, while resolution plays a secondary role. In contrast, Qwen2.5‑VL‑72B exhibits near-constant performance across all combinations, indicating its capacity to flexibly trade temporal for spatial information. These findings suggest that \textbf{smaller models rely more on temporal cues to compensate for limited spatial encoding, whereas larger models can extract complementary signals from both dimensions interchangeably.}

\subsection{Simple Data Synthesis  Approach}

One reason existing models may underperform is that they have never been trained on video image formatted data. To test this hypothesis, we propose a simple data synthesis approach to automatically generate IV‑Bench–aligned examples from existing video QA datasets. Our pipeline involves three primary steps:

\begin{itemize}
\item \textbf{Entity Extraction and Question Rewriting}: Relevant noun entities are automatically extracted from original video QA questions and replaced with image references, reformulating questions to align with the IV-Bench format (e.g., transforming "How many goals did the blue player score in the video?" to "How many goals did the entity in the image score in the video?").

\item \textbf{Image Necessity Filtering}: To ensure image contexts are essential, we employ a "Model-as-Judge" approach using QwQ-32B to filter out queries answerable without images, thus retaining only those genuinely requiring image-based inference.

\item \textbf{Image Extraction}: For the retained queries, corresponding images are automatically extracted from relevant video frames based on entity-frame similarity computed by CLIP-L.
\end{itemize}

Using this approach, we construct approximately 36K synthetic samples aligned with IV-Bench from a subset of the llava-video178k dataset. We fine-tune two variants of llava-onevision-si: one using IV-Bench-aligned synthetic data combined with small amounts of pure video and image data, and another using the original video QA data (prior to alignment) similarly combined with pure video and image data.

We evaluate four models on IV-Bench: the two fine-tuned variants described above, the base model llava-onevision-si before fine-tuning, and the llava-video model fine-tuned using the entire llava-video178k dataset. The results are summarized in Table \ref{tab:model-performance}. Key findings include:

\begin{itemize}
\item Incorporating a small amount of IV-Bench-aligned synthetic data modestly improves performance (from 15.54\% without fine-tuning and 15.23\% with video-only fine-tuning, to 16.13\% with IV-aligned fine-tuning), demonstrating the specific benefit of aligning synthetic data with the IV-Bench format.
\item Scaling with video data also slightly enhances performance (from 15.23\% with limited video data to 16.72\% using extensive video data), indicating incremental gains through increased data volume.
\item Despite applying our synthetic data approach, the observed performance gains were marginal, indicating that mere exposure to video–image–formatted examples is insufficient to close the gap. This outcome underscores the fundamental challenge of image grounded video perception and reasoning in IV‑Bench and suggests that more advanced approaches are required.
\end{itemize}

\begin{table}[htbp]
    \centering
    \caption{\centering Performance of different model variants on IV-Bench.}
    \begin{tabular}{l c}
        \toprule
        \textbf{Model} & \textbf{Accuracy} \\
        \midrule
        LLaVA-OneVision-7B-si & 15.54\% \\
        LLaVA-OneVision-7B-si + 36K IV data & 16.13\% \\
        LLaVA-OneVision-7B-si + 36K Video data & 15.23\% \\
        LLaVA-OneVision-7B-si + LLaVA-Video-178K & 16.72\% \\
        \bottomrule
    \end{tabular}
    \label{tab:model-performance}
\end{table}


\section{Related Work}

\subsection{Multimodal Large Language Models}


Multimodal Large Language Models (MLLMs) have made remarkable progress in recent years. These models typically combine a Large Language Model (LLM) backbone with visual encoders, leveraging visual instruction tuning to enhance their multimodal understanding capabilities \citep{liu2023visual,liu2024llavanext,zhu2023minigpt,ma2024vista,zhang2024flash,qwen2.5-VL}. In the video domain, several models utilize video instruction datasets and employ specialized video projectors to encode videos into visual tokens compatible with LLMs \citep{damonlpsg2024videollama2,2023videochat}. For instance, Video-LLaMA \citep{damonlpsg2023videollama} employs ViT \citep{2020arXiv201011929D} combined with a Q-Former \citep{li2023blip} for frame-level and temporal modeling, whereas LLaMA-Vid \citep{li2024llama} utilizes frame compression techniques to facilitate processing of long videos. Additionally, some MLLMs, such as InternVL2.5 and Qwen2.5-VL, are trained on diverse datasets containing single-image, multi-image, and video data to enhance their adaptability across different modalities. However, our analysis reveals that despite extensive training on diverse datasets, current MLLMs notably lack image-grounded video perception and reasoning data in their training corpora. This makes it particularly valuable to explore the capabilities of MLLMs in this regard. Therefore, we propose IV-Bench, a benchmark explicitly designed to evaluate the capabilities of MLLMs in image-grounded video perception and reasoning.

\subsection{Video Understanding Benchmarks}


Video benchmarks have rapidly evolved, with specialized benchmarks now specifically targeting tasks like temporal perception \citep{2024arXiv240509711W}, action understanding \citep{2023arXiv230510683W,2024arXiv240509711W}, and video reasoning \citep{2021arXiv210508276X}. Recent efforts also include MVBench \citep{li2024mvbench}, a comprehensive short-video benchmark focusing on question-answering to assess general multimodal capabilities, and LongVideoBench \citep{wu2025longvideobench}, which evaluates the reasoning abilities of MLLMs over hour-long videos through a novel referring reasoning task, highlighting significant challenges even for advanced models. Video-MME \citep{fu2024videommefirstevercomprehensiveevaluation} provides a comprehensive benchmark for video analysis, covering durations from 11 seconds up to 1 hour, enabling the assessment of multimodal capabilities, including audio understanding. V2P‑Bench \citep{zhao2025v2p} is a comprehensive benchmark comprising 980 videos and 1,172 visual‑prompt QA pairs across five tasks and twelve fine‑grained dimensions for assessing video understanding with visual prompts. Despite these advances, none of the existing video benchmarks specifically assess image-grounded video perception and reasoning. IV-Bench aims to bridge this research gap.

\section{Conclusion}

In this work, we introduce IV-Bench, the first benchmark explicitly designed to evaluate models on image-grounded video perception and reasoning tasks. Our extensive evaluation of both open-source and closed-source multimodal large language models reveals significant limitations, particularly in effectively leveraging image contexts for accurate video comprehension, notably in temporally sensitive perception and reasoning tasks.  We observe that smaller models demonstrate a limited capability of image grounded video perception and reasoning. Furthermore, our analysis indicates that increasing the number of video frames generally has a more pronounced impact on performance compared to enhancing video resolution. To determine whether the performance gap stems from a lack of video–image formatted data in training, we apply a simple pipeline to convert existing video QA examples into IV‑Bench style. The minimal improvements confirm that the challenges of IV‑Bench extend beyond mere format alignment. We hope IV-Bench will inspire future research to substantially advance the capabilities of multimodal large language models in image-grounded video perception and reasoning.

\clearpage

\section{Contributions and Acknowledgments}

Multimodal Art Projection (M-A-P) is a non-profit open-source AI research community, run by donations.
The community members are working on research topics in a wide range of spectrum, including but not limited to the pre-training paradigm of foundation models, large-scale data collection and processing, and the derived applications on coding, reasoning, and music generation.

\textbf{Leading Authors}
\begin{multicols}{2}
    \begin{itemize}
        \item David Ma, M-A-P
        \item Yuanxing Zhang
        \item Jincheng Ren, M-A-P
        \item Jarvis Guo, M-A-P
    \end{itemize}
\end{multicols}

\textbf{Contributors}
\begin{multicols}{2}
    \begin{itemize}
        \item Yifan Yao, M-A-P
        \item Zhenlin Wei, M-A-P
        \item Zhenzhu Yang, M-A-P
        \item Zhongyuan Peng, M-A-P
        \item Boyu Feng
        \item Jun Ma
        \item Xiao Gu
        \item Zhoufutu Wen, M-A-P
        \item King Zhu, M-A-P
        \item Yancheng He, M-A-P
        \item Meng Cao, MBZUAI
        \item Shiwen Ni, SIAT-CAS
        \item Jiaheng Liu, M-A-P, NJU
        \item Wenhao Huang, M-A-P
    \end{itemize}
\end{multicols}

\textbf{Corresponding Authors}
\begin{multicols}{2}
    \begin{itemize}
        \item Ge Zhang, M-A-P
        \item Xiaojie Jin, M-A-P
    \end{itemize}
\end{multicols}

\newpage

\bibliography{main.bib}

\newpage
\appendix

\newtcolorbox[auto counter, number within=section]{methodbox}[2][]{%
  colback=white, 
  colframe=teal!80!green!80!black,  
  width=\textwidth,
  arc=2mm, 
  boxrule=0.5mm, 
  title={\normalsize\faWrench\hspace{0.5em}#2}, 
  breakable,
  fonttitle=\Large, 
  fontupper=\small, 
  #1
}

\section{Annotation Tutorial}
\label{appendix: annotation tutorial}
\subsection{Question Type}
\begin{figure*}[t]
    \centering
    \includegraphics[width=\textwidth]{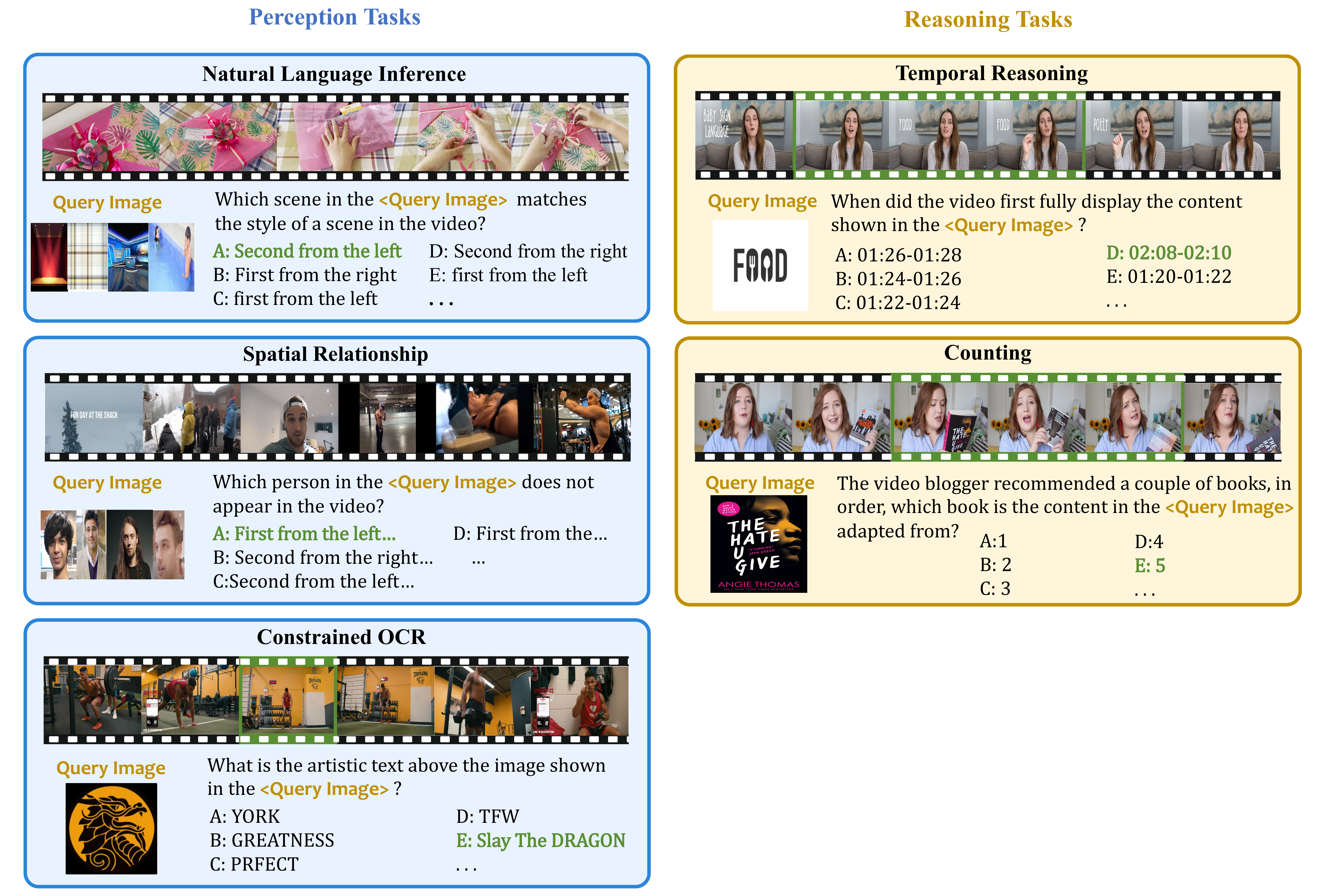}
    \caption{Five remaining IV‑Bench task categories: Natural Language Inference, Constrained OCR, Spatial Relationship, Reasoning, and Temporal Reasoning. Each example requires using text, image, and video together.}
    \label{fig:remaining_tasks}
\end{figure*}

\begin{methodbox}{Question Type Details}

Representative examples of all 13 task categories in IV‑Bench are shown in Figure~\ref{fig:tasks} and Figure~\ref{fig:remaining_tasks}.

\subsection*{Summarization Questions}
\begin{itemize}
\item These questions aim to test the model's ability, using the provided image, to understand the main narrative thread of the video and summarize the main plot or character stories from a global perspective. The model needs to leverage clues or context from the image, go beyond understanding single frames or segments, grasp the core content of the video, and comprehend and extract a plot summary.
\end{itemize}
\subsection*{Spatial Relationship Questions}
\begin{itemize}
\item These questions aim to test the model's ability, using the provided image, to understand the spatial positions of objects and their relationships within the video scene. The model needs to identify specific objects or areas based on the image, locate them in the video scene, and describe the surrounding spatial layout and relationships with other objects.
\end{itemize}
\subsection*{Existence Questions}
\begin{itemize}
\item These questions aim to test the model's ability, using the provided image, to identify a specific object within the image and search within the video content to determine if that object appears or is used in the video.
\end{itemize}
\subsection*{Reverse Existence Questions}
\begin{itemize}
\item These questions aim to test the model's ability, using the provided image (often showing a set), to perform a comparative analysis of the set and identify missing elements within the video content. The model needs to identify all items in the image set, compare them against the video content, and identify those items that do not appear in the video.
\end{itemize}
\subsection*{Natural Language Inference Questions (NLI)}
\begin{itemize}
\item These questions aim to test the model's ability, using the provided image, to perform consistency reasoning between the visual content of the image and the video. The model needs to understand the image's visual information and the video's content to determine if the image is semantically consistent with the video content.
\end{itemize}
\subsection*{Detailed Events Questions}
\begin{itemize}
\item These questions aim to test the model's ability, using the provided image, to identify a specific object or scene in the image, locate related events within the video, and extract specific detail information (e.g., price, time, location) from those events.
\end{itemize}
\subsection*{Explanation/Instruction Questions}
\begin{itemize}
\item These questions aim to test the model's ability, using the provided image, to understand the attributes of the object shown in the image and perform an associative analysis connecting them with explanatory content in the video (such as introductions or descriptions). Based on the video content, the model needs to understand the reason, definition, function, impact, or creation process related to specific attributes of the object shown in the image.
\end{itemize}
\subsection*{Keyframe Extraction Questions}
\begin{itemize}
\item These questions aim to test the model's ability to integrate textual understanding with visual analysis of the provided image to identify a specific object or state. The model must then locate the corresponding keyframe(s) or segment within the video timeline, as depicted in the image, demonstrating comprehension of the 'keyframe' concept and the ability to correlate visual cues with temporal positioning.
\end{itemize}
\subsection*{Counting Questions}
\begin{itemize}
\item These questions aim to test the model's ability, using the provided image, to identify the specific object category designated in the image and count all instances of that object within the video scene(s). The model needs to accurately identify and differentiate between individual instances and report the total count.
\end{itemize}
\subsection*{Spatiotemporal Calculation Questions - Spatial Dimension}
\begin{itemize}
\item These questions aim to test the model's ability, using the provided image (which might include scale information, a map, or specific reference objects), to understand spatial scales and track the movement trajectory of objects/people within the video, ultimately calculating the actual distance traveled. The model needs to utilize the scale or reference points from the image to analyze the motion depicted in the video.
\end{itemize}
\subsection*{Spatiotemporal Calculation Questions - Temporal Dimension}
\begin{itemize}
\item These questions aim to test the model's ability, using the provided image (which might reference specific time points, events, or individuals), to understand and analyze temporal information and sequential events within the video. The model needs to use the clues from the image to perform temporal calculations (like duration), comparisons (such as length), or pinpoint events at specific times within the video.
\end{itemize}
\subsection*{Limited OCR Questions}
\begin{itemize}
\item These questions aim to test the model's text recognition capabilities under specific constraints (like artistic lettering or particular fonts), using a provided image that showcases a specific text style or example. The task involves searching for and extracting text content from the video that matches the specified style.
\end{itemize}
\subsection*{Attribute Change Questions}
\begin{itemize}
\item These questions aim to test the model's ability, using the provided image (which typically designates the object/person to track), to continuously follow that specific target across different segments of the video and analyze/describe how its attributes (e.g., color, state, location) change over time.
\end{itemize}
\subsection*{Temporal Reasoning Questions}
\begin{itemize}
\item These questions aim to test the model's ability, using the provided image (which might reference event types, participants, or scenes), to understand the sequence of recurring events within the video and perform temporal reasoning to locate the specific time point or time interval corresponding to the Nth occurrence of a particular event in that sequence.
\end{itemize}
\end{methodbox}

\subsection{Data Annotation Steps}
\begin{methodbox}{Operations}

\subsection*{Watch Video and Determine Question Type:}
\begin{itemize}
\item After watching the video content to be annotated in its entirety, select the most suitable and valuable question type from the 14 pre-defined types based on the video content, and then brainstorm the question direction accordingly to prepare for subsequent question stem and answer design.
\end{itemize}

\subsection*{Design Question Stem and Answer:}
\begin{itemize}
\item \textbf{Question Stem Design Requirements:}
\begin{itemize}
 \item \textbf{Close Relevance:} Ensure the question stem is closely related to the content of the video and paired images.
\item \textbf{Information Confidentiality:} The question stem should avoid revealing any direct information about the video and images, retaining only necessary hints.
\item \textbf{Assessment Significance:} Question stem design should have assessment significance, avoiding simple questions of purely objective facts.
\item \textbf{Concise and Clear Language:} Use concise and clear language to describe the question, avoiding ambiguity or redundancy.
\end{itemize}
\subsection*{Answer Design Requirements:}

\begin{itemize}
\item \textbf{Unique Clarity:} For the question stem, there must be a unique and clearly correct answer.
\item \textbf{Information Confidentiality:} The content of the correct answer itself must not directly reveal any information about the paired images.
\item \textbf{Video Granularity:} Specify the smallest video unit required to answer the question, such as:
\begin{itemize}
\item \textbf{Frame:} The answer is located at a specific frame in the video.
\item \textbf{Clip:} The answer is located in a continuous clip of the video.
\item \textbf{Full Video:} The answer requires understanding the full video content.
\end{itemize}
\item \textbf{Video Range:} Clearly indicate the specific segment range in the video where the answer is located to quickly verify the accuracy of the answer.
\end{itemize}
\end{itemize}

\subsection*{Collect Paired Images:}

\begin{itemize}
\item \textbf{Diversity of Image Sources:} Widely collect images that meet the question requirements from the internet or video resources, avoiding single and over-reused image sources to ensure image diversity.
\item \textbf{Non-Video Screenshots:} Directly capturing frames from the current test video as paired images is prohibited.
\end{itemize}

\subsection*{Image Quality Assurance:}

\begin{itemize}
\item \textbf{Texture Clarity:} The texture of the image must be clearly distinguishable, avoiding blurriness to ensure the effectiveness of visual information.
\item \textbf{Subject Consistency:} The characters or objects in the image must be completely consistent with or visually highly similar in texture to the characters or objects appearing in the video.
\item \textbf{Subject Prominence:} The target object in the image should occupy the main position of the image, highlighting the subject and reducing background interference for easy observation and identification.
\item \textbf{Visual Information Validity:} Images must meet pre-defined "visual information validity requirements."
\item \textbf{Annotation Validity Requirement Number:} Based on the specific basis for selecting images, annotate the corresponding number (requirement\_number) of the "visual information validity requirements" that the image meets. (Please provide a pre-defined list of "visual information validity requirement" numbers for accurate annotation).
\end{itemize}

\subsection*{Design Distractor Options (9 Options):}

\begin{itemize}
\item \textbf{Number of Distractor Options:} Each question needs to design 9 misleading incorrect options, plus 1 correct answer, for a total of 10 options.
\item \textbf{Diversified Confusion Strategies:} Avoid patterned option design. Incorrect answers should be set from different angles and dimensions to increase the discrimination and difficulty of the questions. 
Common confusion strategies include:
\begin{itemize}
\item \textbf{Conceptual Confusion:} Use options with concepts similar to the correct answer but with subtle differences in meaning to test the precise understanding of concepts.
\item \textbf{Partial Correctness:} The content described in the option partially matches the video content, but does not fully answer the question raised in the question stem.
\item \textbf{Irrelevant Information:} The option content has low direct relevance to the video content, but may have some connection with the question in daily cognition or common sense, setting up interference.
\item \textbf{Incorrect Reasoning:} Conclusions obtained from logical reasoning that seems reasonable but is actually incorrect based on the video content are used as distractor options.
\item \textbf{Image Misdirection:} Use objective facts or visual information presented in paired images to design options that match the images but not the video, forming interference.
\item \textbf{Conclusion Divergence:} For the same question stem, change different paired images to make the conclusion of the options change with the images, and use this as a distractor item.
\item \textbf{Real-World Significance Trap:} Design options that have certain significance or rationality in real life based on the images, but do not conform to the video content, inducing users to answer based on common sense rather than the video.
\item \textbf{Avoid Fabrication:} Avoid designing options with overly obvious traces of fabrication, ensuring that distractors have a certain degree of misleadingness and avoiding easy exclusion by users.
\end{itemize}
\end{itemize}

\subsection*{Option Format Requirements:}

\begin{itemize}
\item \textbf{Length Consistency:} Ensure that the lengths of the 10 options (including the correct answer and 9 distractors) are approximately equal to avoid answer information leakage due to option length differences.
\item \textbf{Concise Language:} The language expression of distractor options should be concise and clear, avoiding unnecessary complex sentence structures and rare vocabulary. At the same time, the descriptive language of distractor options should avoid revealing any information about the paired images.
\end{itemize}
\end{methodbox}

\subsubsection{Methods for Designing Distractors}
\begin{methodbox}{Distractors Design Methods}
\begin{itemize}
    \item \textbf{Visual Replacement:} Replace a visual element in the video (such as the color, shape, or texture of an item) with visual information that is similar but inaccurate to the actual content.
    \item \textbf{Quantitative Replacement:} Replace a numerical detail in the video (such as quantity, time, distance, etc.) with an incorrect numerical value.
    \item \textbf{Spatial Replacement:} Incorrectly describe the location where an event occurs, for example, misdescribing "in the kitchen" as "in the living room" or another space.
    \item \textbf{Temporal Replacement:} Incorrectly describe the time point when an event occurs, for example, misdescribing "morning" as "evening."
    \item \textbf{Addition of Information:} Deliberately add non-existent events or information to the video content, such as fabricating a plot or detail.
    \item \textbf{Missing Information:} Delete important information or events that exist in the video, for example, deliberately omitting key information, leading to incomplete information.
    \item \textbf{Detail Replacement:} Incorrectly replace key information involving characters, events, or details in the video, for example, replacing the profession or age attributes of a character, or incorrectly describing the details of an event.
    \item \textbf{Sequential Replacement:} Arrange a series of actions or events that actually occurred in the video in the wrong order, disrupting their chronological relationship.
    \item \textbf{Reality Trap:} Based on people's common sense or logic in real life, design options that have a certain meaning or rationality in reality, but these options do not match the actual content of the video.
    \item \textbf{Conclusion Divergence:} For the same video content and question stem, by changing different paired images, the conclusion of the options changes with the paired images.
\end{itemize}
\end{methodbox}

\begin{methodbox}{Video Content Guidelines and Question Screening Process}

\textbf{Video Content Guidelines:}
\begin{itemize}
    \item \textbf{Include Similar Distractors:}
    \begin{itemize}
        \item The video must include at least two figures or objects of the same type as the image target to create visual confusion.
    \end{itemize}

    \item \textbf{Meet any of the following conditions to emphasize the importance of texture:}
    \begin{itemize}
        \item \textbf{Condition 1: Texture-Dominant Definition}
        \begin{itemize}
            \item The image target can only be identified in the video through texture characteristics.
            \item \textit{For example: Only present close-up texture details of a human face, weakening features like contours and clothing.}
        \end{itemize}

        \item \textbf{Condition 2: Key Feature Differentiation}
        \begin{itemize}
            \item Distractors of the same type in the video differ from the image target in at least one key visual feature.
            \item \textit{For example: Shoes of the same style but different colors, the same person wearing different styles of clothing.}
        \end{itemize}

        \item \textbf{Condition 3: Multiple Feature Description Requirement}
        \begin{itemize}
            \item The image target requires at least four visual features to be fully described.
            \item Emphasize the complexity of the target's visual information, requiring multi-dimensional features for accurate understanding.
        \end{itemize}
    \end{itemize}
\end{itemize}

\vspace{0.5em}

\textbf{Question Screening Process:}
\begin{itemize}
    \item \textbf{Generate Detailed Description:}
    \begin{itemize}
        \item After annotation, use MLLM to generate a Detailed Description based on the image and question, controlling the granularity of the text description.
    \end{itemize}

    \item \textbf{Assess Answerability:}
    \begin{itemize}
        \item Use MLLM or human evaluation, combined with the Detailed Description + video, to attempt to answer the question.
    \end{itemize}

    \item \textbf{Question Screening:}
    \begin{itemize}
        \item Eliminate questions that can be answered correctly based solely on the Detailed Description + video.
        \item Retain questions to ensure that image texture information is crucial for a correct answer.
    \end{itemize}
\end{itemize}

\end{methodbox}


\newtcolorbox[auto counter, number within=section]{purposebox}[2][]{%
  colback=white, 
  colframe=blue!50!black, 
  width=\textwidth,
  arc=2mm, 
  boxrule=0.5mm, 
  title={\normalsize\faCompass\hspace{0.5em}#2},
  breakable, 
  fonttitle=\Large, 
  fontupper=\small, 
  drop shadow southeast, 
  #1
}

\newtcolorbox[auto counter, number within=section]{promptbox}[2][]{%
  colback=white, 
  colframe=purple!70!blue!80!black,  
  width=\textwidth,
  arc=2mm, 
  boxrule=0.5mm, 
  title={\normalsize\faInfoCircle\hspace{0.5em}#2},
  breakable,
  fonttitle=\Large, 
  fontupper=\small, 
  drop shadow southeast, 
  #1
}
\section{Quality Control Process Details}
The quality inspection of IV-Benchmark comprises two rounds:
\begin{itemize}
    \item \textbf{Round 1:} focuses on standardizing problem structures and content validity.
    \item \textbf{Round 2:} addresses advanced quality requirements to ensure task rigor.
\end{itemize}

\subsection{Round 1 Quality Control}

\begin{purposebox}{Purpose}
\begin{itemize}
     \item Ensure basic structural integrity and content standardization, including unambiguous question formulation, verifiable answers, reasonable distractors, and data quality.
\end{itemize}
\end{purposebox}

\begin{promptbox}{Quality Assessment Dimensions}
\begin{itemize}
    \item \textbf{Clarity Validation:} Verify grammatical correctness and unambiguous expression of questions/options.
    \item \textbf{Answer Validity Validation:} Confirm answers are deducible from video content (eliminate labeling errors).
    \item \textbf{Task Categorization Calibration:} Validate proper classification of question types.
    \item\textbf{Contextual Validation :} Contextual Validation: Ensuring answers and distractors are plausible and contextually relevant.
    \item \textbf{Image Quality Assurance:} Check query image resolution and visibility of critical information.
    \item \textbf{Option Completeness Validation:} Verify coverage of plausible alternatives (e.g., critical missing options in multi-choice questions).

\end{itemize}
\end{promptbox}

\subsection{Round 2 Quality Control}
\begin{purposebox}{Purpose}
\begin{itemize}
    \item Ensure task validity by verifying the necessity of multimodal components and the contextual plausibility of distractors, thereby mitigating evaluation bias caused by design flaws.
\end{itemize}
\end{purposebox}

\begin{promptbox}{Methods}
\begin{itemize}
    \item \textbf{Multimodal Necessity Check:} Retain only questions requiring combined analysis of text/image/video.
    \item \textbf{Information Leakage Detection:} Identify text queries that inadvertently reveal visual content through textual cues (e.g., explicit object descriptions, positional references) and eliminate leakage by rewriting queries to preserve task intent.
    \item \textbf{Commonsense Dependency Screening:} Eliminate questions answerable through general knowledge alone (e.g., "the sun rises in the east").
    \item \textbf{Distractor Optimization:} Redesign meaningless distractors based on video content. Preferred: Distractors should match answer categories or create confusion (e.g., use actual OCR text from videos for text-related questions).
      \item \textbf{Object Uniqueness Verification:} Ensure non-unique targets in questions (e.g., "What color is the person's clothing shown in both image and video?" requires multiple persons in video).
\end{itemize}
\end{promptbox}

\begin{figure*}[t]
    \centering
    \includegraphics[width=\textwidth]{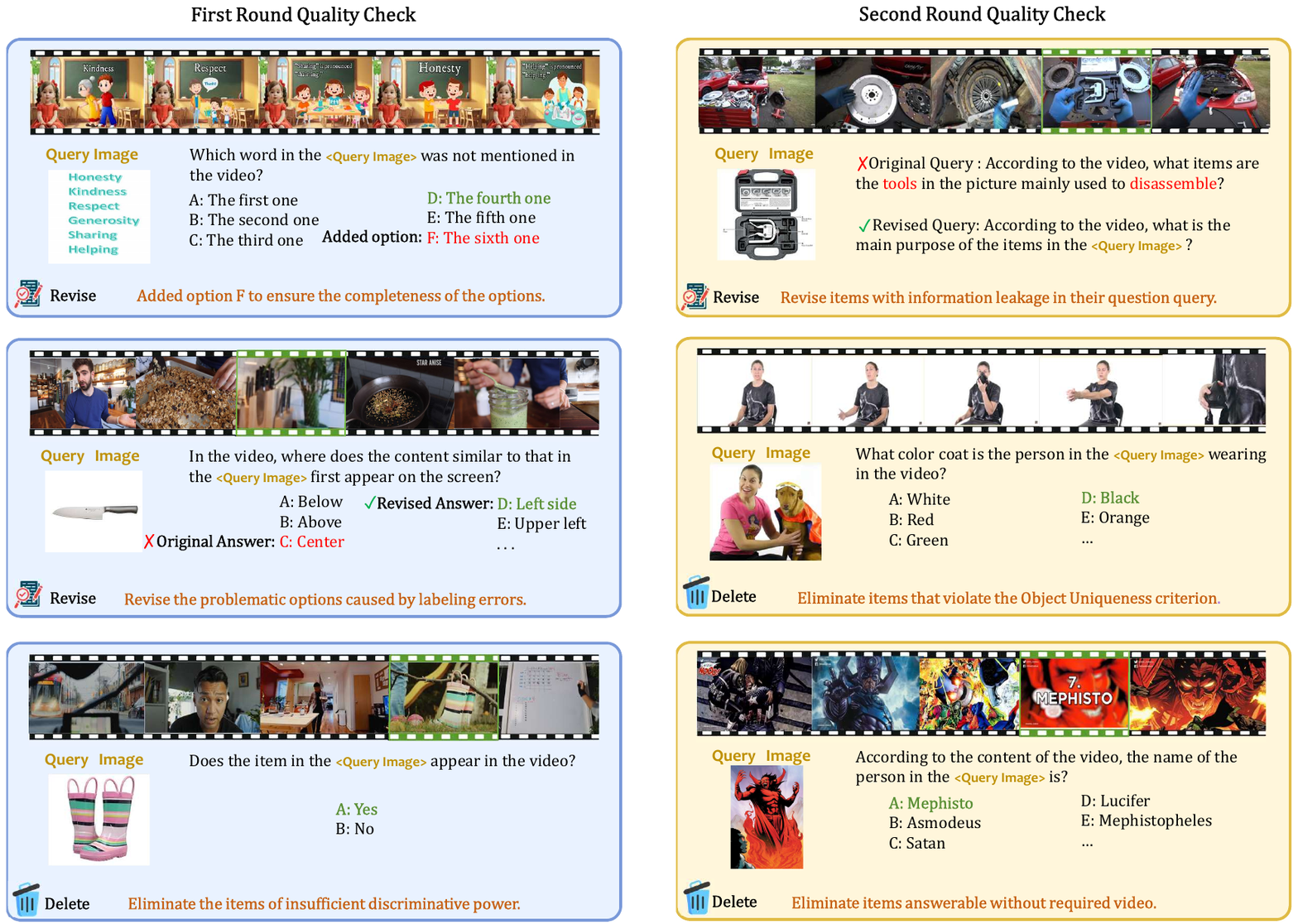}
    \caption{Representative data examples from Two Round Quality Check. Each round includes modifications and deletions to queries and images.}
    \label{fig:tasks_qc}
\end{figure*}

\section{Infernce prompt}

Multiple-choice questions are constructed by pairing each instance with one correct answer and several distractors. During inference, the answer choices are randomly shuffled to ensure that the correct answer appears in different positions. The default placement order is video, image, and text. For some models, we also experiment with placing the image after the video. Therefore, we have two different text prompts: one with the order of video, image, and text, and another with the image placed after the video. In the prompt, we explicitly specify the order of the video and image, the video length, and instruct the model to answer the question based on both the video and the image, providing only the options.
\begin{promptbox}{Prompt}
\begin{itemize}
    \item \textbf{Video-first prompt:} $<$VIDEO$>$ $<$IMAGE$>$ We provide you with an image placed at the very beginning, followed by a video that has been divided into \texttt{evenly spaced frames} across its \texttt{seconds duration}. Please answer the question based on the content from both the image and the extracted video frames.
    \item \textbf{Image-first prompt:} $<$IMAGE$>$ $<$VIDEO$>$ We provide you with a video that has been divided into \texttt{frame\_num} evenly spaced frames across its \texttt{duration} seconds duration, followed by an image. Please answer the question based on the content from both the video frames and the image.
\end{itemize}
\end{promptbox}
\section{Comparison with Video‑MMMU}

\begin{figure*}[t]
    \centering
    \includegraphics[width=\textwidth]{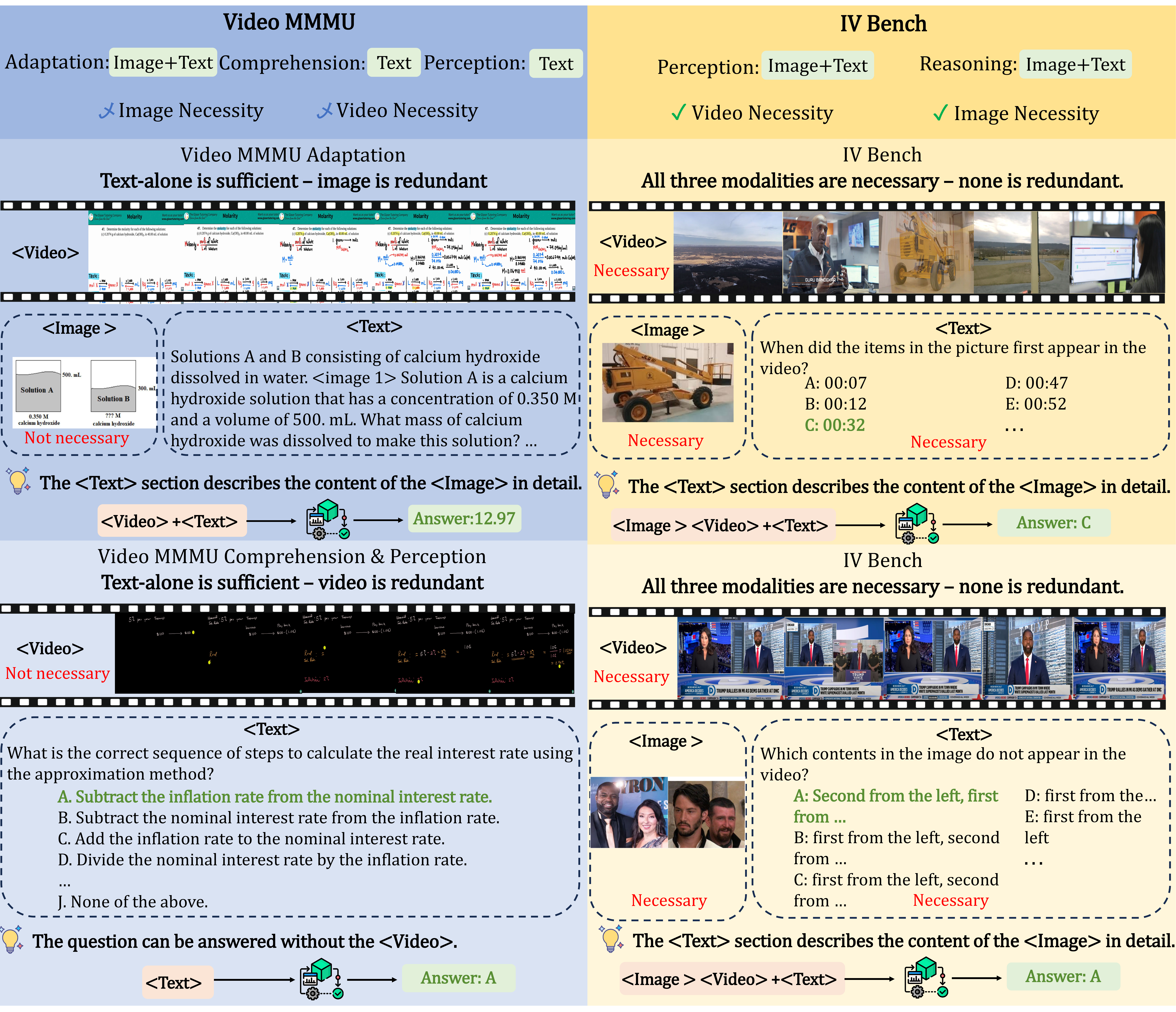}
    \caption{Side-by-side comparison of Video-MMMU (left) and IV-Bench (right). The left side shows that in Video-MMMU many questions can be answered using only text, so the image or video input is often not needed. The right side shows that IV-Bench always requires using the video, image, and text together to answer each question. IV-Bench enforces true multimodal reasoning by adding visual distractors and making each modality necessary, so no single modality (like text alone) can answer the questions.}
    \label{fig:comparsion_app}
\end{figure*}

Figure~\ref{fig:comparsion_app} shows the main differences between IV‑Bench and Video‑MMMU. IV‑Bench is designed to require all three modalities—text, image, and video—for every question. In contrast, Video‑MMMU often allows questions to be answered using only text, making video or image redundant. Specifically:

\textbf{Image Necessity}:In Video‑MMMU, image‑text prompts often include full descriptions that make the image redundant. For instance, the calcium‑hydroxide question’s text describes both solutions in detail, so viewing the picture of the bottles does not change the answer. In IV‑Bench, each image contains distractor objects that are not mentioned in the text. Only by examining the image can one distinguish these distractors—e.g.\ in the "Which contents…" question, you must look at the picture to see which item never appears in the video.

\textbf{Video Necessity}: In Video‑MMMU (left of Fig.\ref{fig:comparsion_app}), the real‑interest‑rate question includes both text and video, but it can be answered using only the text, making the video unnecessary. In fact, in the very first Video‑MMMU example, neither the image nor the video is needed to find the correct answer. By contrast, in IV‑Bench (right of Fig.\ref{fig:comparsion_app}), the question "When did the items in the picture first appear in the video?” cannot be answered without watching the video, since only the video reveals the exact timestamp. 

\textbf{Query Modality}: Video‑MMMU uses pure‑text for two‑thirds of its items: in the Fig.\ref{fig:comparsion_app} "Comprehension" and "Perception" examples (the second row), both queries are text‑only, accounting for 2/3 of the dataset. IV‑Bench, however, pairs text with an image in every query. For example, in the "Which contents in the image do not appear in the video" item, the model must read the text prompt and inspect the image together to identify the correct distractor.

\end{CJK*}
\end{document}


\begin{CJK}{UTF8}{gbsn}
\begin{longtable}{
  >{\raggedright\arraybackslash}p{8cm}|  
  >{\raggedright\arraybackslash}p{8cm}   
}
  \caption{Textbook List for Question Source Reference} \\
  \toprule
  \endfirsthead
  \multicolumn{2}{c}{Continued} \\
  \toprule
  \endhead
  \bottomrule
  \endfoot
  \bottomrule
  \endlastfoot

  \rowcolor{science}
  \multicolumn{2}{c}{\cellcolor{science}\textbf{SCIENCE}} \\
  \midrule
   \textit{Tools of Radio Astronomy} &  \textit{Fusion Plasma Physics}\\
   \textit{Astronomy Today} & \textit{Materials Science and Engineering, An Introduction} \\
   \textit{Introduction to Astronomy and Cosmology} & \textit{Black Holes, White Dwarfs, and Neutron Stars} \\
   \textit{Abstract Algebra: Theory And Applications} &\textit{Introduction To Physical Oceanography}\\
   \textit{Essentials Of Geology} & \textit{Atoms, Radiation, And Radiation Protection} \\
   \textit{Physical Principles Of Sedimentology: A Readable Textbook For Beginners And Experts} & \textit{Theoretical Physics Text and Exercise Books: Volume 5: Gauge Theory of Weak Interactions} \\
   \textit{Textbook Of Engineering Geology} & \textit{Introduction To Mathematical Physics} \\
   \textit{The Changing Earth: Exploring Geology And Evolution} & \textit{Quantum Paradoxes: Quantum Theory For The Perplexed} \\
   \textit{Elements Of Physical Oceanography} &\textit{Physics Of Sedimentology: Textbook And Reference} \\
   \textit{Entangled Systems: New Directions In Quantum Physics} & \textit{Rigid Body Mechanics: Mathematics, Physics And Applications}  \\
   \textit{Introduction To Thermodynamics And Kinetic Theory Of Matter} &   \textit{Ultrahigh Pressure Metamorphism: University Textbook} \\
   \textit{Introduction To Ocean Circulation And Modeling} & \textit{Inverse Methods In Physical Oceanography} \\
   \textit{Nonlinear Physical Oceanography} & \textit{Physical Oceanography Of Coastal And Shelf Seas} \\
  《高分子化学与物理学习指导及习题集》 & 《放射化学习题汇编》 \\
  《物理化学学习指导》 &  《随机过程及应用习题集》\\
  《基础有机化学》 & 《声学习题集》 \\
  《物理化学解题思路和方法》 & 《天体物理概论》 \\
  《粒子天体物理》 & 《无机化学》 \\
  《遗传学学习指导与题解》 & 《半导体物理学 学习题集及详解》 \\
  《半导体物理习题及解答》 & 《天体测量和天体力学基础》 \\
  《细胞生物学》 & 《近代物理习题解答》 \\
  《热学习题思考题解题指导》 & 《细胞生物学实验与习题指导》 \\
  《原子(近代)物理学科学化考试理论与实践》 & 《化学 中心科学》 \\
  《美国物理试题与解答 第4卷 原子物理学、核与粒子物理学》 & 《美国物理试题与解答 第五卷: 热力学与统计物理学》 \\
  《概率与随机过程习题集 下 随机过程》 & 《随机过程习题解析》 \\
  《固体物理导论》 & 《高等学校理工类课程学习辅导丛书》 \\
  《Kittel固体物理导论习题详解》 & 《物理学大题典 相对论物理学》 \\
  《物理学大题典 原子分子物理学》 & 《水声学原理》 \\
  《微分几何》 & 《模糊数学原理及应用》 \\
  《数值分析学习辅导 · 习题解析》 & 《固体物理学》 \\
  《综合化学》 & 《电磁学》\\
  《光学》 & 《离散数学及其应用》 \\
  《现代分子生物学》 & 《物理化学》 \\
  《概率论》 & 《数理统计》 \\
  《物理化学》 & 《遗传学学习指导与题解》 \\
  《高分子化学》 & 《生物化学》\\
  《细胞生物学》 & 《现代分子生物学》 \\
  《数学分析教程》 & 《吉米多维奇数学分析习题集题解》\\

\rowcolor{engineer}
\multicolumn{2}{c}{\cellcolor{engineer}\textbf{ENGINEERING}} \\
\midrule
    \textit{Fundamentals of Ship Hydrodynamics: Fluid Mechanics, Ship Resistance and Propulsion} & \textit{Process Control Engineering: A Textbook for Chemical, Mechanical and Electrical Engineers} \\
    \textit{Textbook of Seismic Design: Structures, Piping Systems, and Components} & \textit{Simulation and Optimization of Furnaces and Kilns for Nonferrous Metallurgical Engineering} \\
    \textit{The Principles of Naval Architecture Series: Strength of Ships and Ocean Structures} & \textit{Process Software and Digital Networks: Instrument Engineers’ Handbook}  \\
    \textit{Unconventional Oil and Gas Resources: Exploitation and Development} & \textit{Production and Transport of Oil and Gas: Gathering and Transportation}\\
    \textit{Radioactive Investigations of Oil and Gas Wells}  & \textit{Residential Land Development Practices} \\
    \textit{Radio Navigation Systems for Airports and Airways} & \textit{Fundamentals of Solid-State Electronics}  \\
    \textit{Instrument Engineers' Handbook: Process Control and Optimization} & \textit{A Textbook of Fluid Mechanics and Hydraulic Machines}\\
    \textit{Fluid Mechanics for Engineers: A Graduate Textbook} &   \textit{Materials Science and Engineering: An Introduction} \\
   \textit{Fundamentals of Drilling Engineering}& \textit{A Textbook of Production Technology} \\
    \textit{Sea Loads on Ships and Offshore Structures} & \textit{A Textbook of Thermal Engineering} \\
    \textit{Strength of Ships and Ocean Structures} & \textit{A Textbook on Heat Transfer} \\
    \textit{Process and Applications in Agricultural Engineering} & \textit{Think Java: How to Think Like a Computer Scientist} \\
    \textit{RF Linear Accelerators}  & \textit{Fiber Optic Sensors} \\
    \textit{Fusion: An Introduction to the Physics and Technology of Magnetic Confinement Fusion} & \textit{Phase Diagrams in Metallurgy: Their Development and Application}\\
    \textit{Elementary Engineering Hydrology} & \textit{Transportation Engineering} \\
    \textit{Risk Analysis for Prevention of Hazardous Situations in Petroleum and Natural Gas Engineering} & \textit{Optical Imaging and Photography: Introduction to Science and Technology of Optics, Sensors and Systems}\\
    \textit{Fundamentals of Ship Hydrodynamics} & \textit{Corrosion Engineering: Principles and Practice} \\
    \textit{Introductory Soil and Water Conservation Engineering} & \textit{Prism and Lens Making: A Textbook for Optical Glassworkers} \\
    \textit{Alternating Current Circuits} & \textit{Textbook of Engineering Drawing} \\
    \textit{Food Science and Technology} & \textit{Textbook of Food Science and Technology} \\
    \textit{A Textbook of Fluid Mechanics and Hydraulic Machines}  & \textit{Aquatic Chemistry: For Water and Wastewater Treatment Applications} \\
    \textit{Hydraulic Power Plants} & \textit{Nuclear and Radiochemistry} \\
    \textit{Physiological Ecology of Forest Production} & \textit{A Textbook of Robotics: Basic Concepts}\\
    \textit{Metallurgy Fundamentals} & \textit{Random Signal Analysis} \\
    \textit{Mechanical Metallurgy} \\
        
    《信息论与编码理论:剑桥大学真题精解》 & 《水力学习题解析-上册》 \\
    《水力学习题解析-下册》 & 《水力学学习指导与习题解答》 \\
    《热力学与统计物理学学习指导》 & 《误差理论与测量平差基础习题集》 \\
    《光电大地测量仪器学 习题、实验、检测、维修及操作部分》& 《工程测量习题集》 \\
    《电子线路习题集》 &  《电子技术基础习题集及解答》\\
    《电子技术基础模拟部分第六版 学习辅导与习题解答》& 《电子电路与系统基础(清华大学电子工程系核心课系列教材)》 \\
    《理论力学习题集》 & 《化工流体流动和传热》 \\
    《化工原理 流体流动与传热分册学习指导书》 & 《工程热力学习题解答》 \\
    《无线电通信系统 理论·例题·习题》 & 《电力电子技术学习指导习题集及仿真》 \\
    《流体力学理论例题与习题》 & 《传热学第五版》 \\
    《交通运输工程学》 & 《交通工程学计算示例》 \\
    《交通工程学基础》 & 《高电压绝缘技术》 \\
    《光电技术》 & 《土力学及基础工程学习辅导与习题精解》 \\
    《核反应堆物理分析》 & 《电化学原理》 \\
    《土力学学习指导与考题精解》 & 《无线电通信系统 理论·例题·习题》 \\
    《天线原理习题集》 & 《光纤通信系统》 \\
    《固态电子论》 & 《模拟电路及其应用》 \\
    《物理奥林匹克竞赛大题典(电磁学卷)》 & 《电磁学千题解》 \\
    《运筹学习题集 第5版》 & 《营养与食品卫生学学习指导及习题集》 \\
    《农业经济管理》 & 《电力电子技术习题解答、实验与课程设计指导》 \\
    《公路工程管理与实务命题点全面解读》 & 《城市规划与设计》 \\
    《矿业工程管理与实务复习题集》 & 《模式识别》 \\
    《计算机操作系统学习指导与题解》 & 《岩土工程数值计算》 \\
    《工程材料学》 & 《统计力学》 \\
    《材料固体力学》 & 《农业水土工程必会100题》 \\
    《化学反应工程原理例题与习题》 & 《制冷与低温原理》 \\
    《车辆工程考研题》 & 《材料科学基础》 \\
    《空气动力学基础》 & 《流体力学习题集》 \\
    《工程热力学学习辅导与习题解答》 & 《2000工程热力学习题精解》 \\
    《电磁学第二版习题分析与解答》 & 《物理学大题典 热学 热力学 统计物理 第2版》 \\
    《物理学大题典 量子力学》 & 《轮机工程基础》 \\
    《大地测量学基础》 & 《船舶结构力学习题集》 \\
    《船舶货运习题集》 & 《核技术应用辐射安全与防护》 \\
    《材料科学基础辅导与习题》 & 《结构力学学习指导与典型例题解析》 \\
    《内燃机原理习题集》 & 《染料化学》 \\
    《化工系统工程》 & 《现代控制工程》 \\
    《高电压技术》 & 《应用模糊数学》 \\
    《电化学方法原理和应用》 & 《电力系统分析中的计算方法》 \\
    《结构力学教程》 & 《材料科学基础》\\
    《交通运输工程导论》（第三版） & 《核反应堆物理分析》） \\
    《自动控制理论与设计》 & 《电路基础》（第二版） \\
    《信号与系统》 & 《数字信号处理》（第3版） \\
    《材料力学》& 《化工原理》 \\
    《电动力学》 & 《热力学与统计物理学》 \\
    《机械设计基础》 & 《电力系统分析》 \\
    《电力系统设备与接线》 & 《过程控制系统与仪表》 \\
    《工程材料》 & 《集成电路设计基础》 \\
    《光纤通信系统学习指导与习题解析》 & \\
    
\rowcolor{medical}
\multicolumn{2}{c}{\cellcolor{medical}\textbf{MEDICINE}} \\
\midrule

\textit{A Textbook Of Clinical Pharmacy Practice: Essential Concepts And Skills} & \textit{Drugs For The Heart: Textbook With Online Updates} \\
\textit{Goodman And Gilman'S Manual Of Pharmacology And Therapeutics} & \textit{Pharmaceutical Analysis: A Textbook For Pharmacy Students And Pharmaceutical Chemists Volume 4} \\
\textit{Pharmaceutical Analysis: A Textbook For Pharmacy Students And Pharmaceutical Chemists} & \textit{Pharmaceutical Chemistry 2: Drugs And Their Biological Targets} \\
\textit{Pharmaceutical Chemistry. Volume 2: Drugs And Their Biological Targets 2} & \textit{Pharmaceutical Chemistry: Volume 2 Drugs And Their Biological Targets} \\
\textit{Textbook Of Drug Design And Discovery, Third Edition} &  \textit{The Pancreas : An Integrated Textbook Of Basic Science, Medicine, And Surgery}\\
\textit{The Esc Textbook Of Cardiovascular Medicine} & \textit{Public Health Medicine For The Tropics} \\
\textit{Textbook Of Clinical Occupational And Environmental Medicine} & \textit{Medical Entomology: A Textbook On Public Health And Veterinary Problems Caused By Arthropods} \\
\textit{Textbook Of Hepatology: From Basic Science To Clinical Practice, 2 Volume In One File, 3Rd Ed Two Volumes In One File} & \textit{Textbook Of Drug Design And Discovery} \\
 《医学应试题库丛书 妇产科学》 & 《医学知识记忆与考试一点通 病理学》 \\
 《放射医学》 & 《医学微生物学与寄生虫学学习指导及习题集》 \\
 《生理学》 & 《生物化学》 \\
 《病理学》 & 《内科学》 \\
 《卫生统计学》 & 《流行病学》 \\
 《毒理学基础》 & 《医学免疫学习题集》 \\
 《法医学》 & 《外科学》 \\
 《眼科学》 & 《临床检验基础》 \\
 《流行病学学习指导与习题集》 & 《环境卫生学习题集》 \\
 《药物化学》 & 《药物化学学习指导与习题集》 \\
 《内科主治医师资格考试习题精选集》 & 《预防医学习题精选》 \\
 《全科医学习题集》 & 《检验医学习题集》 \\
 《药剂学学习指导与习题集》 & 《外科学习题集》 \\
 《皮肤性病学习题集》 & 《眼科学习题集》 \\
 《临床医学应试习题集》 & 《免疫学与病原生物学习题集》 \\
 《儿童少年卫生学学习指导与习题集》 & 《药理学应试习题集》 \\
 《免疫学基础与病原生物学习题集》 & 《中医基础理论》 \\
 《口腔内科学》 & 《神经病学》 \\
 《药剂学》 & 《药理学》 \\
 《运动生理学》&《中药学》 \\
 《中医基础理论习题集》&《药物分析》\\
 《药物分析》 & 《药物化学》 \\
  《药剂学》 & 《药理学》 \\
  《运动生理学》 & 《医学微生物学与寄生虫学学习指导及习题集》 \\

  \rowcolor{literature}
  \multicolumn{2}{c}{\cellcolor{literature}\textbf{LITERATURE \& ARTS}} \\
  \midrule

  Translation Between English And Arabic: A Textbook For Translation Students And Educators  & Fundamentals, Function, And Form Theory And Analysis Of Tonal Western Art Music \\
  Excursions In World Music, Seventh Edition Volume 2 \\
  《文献学概要》 & 《西方音乐史》 \\
  《传播学原理》（第三版） & 《跨文化交际》 \\
  《文学理论》 & 《中国文学理论批评史》 \\
  《中国现代文学三十年》 & 《外国文学史》 \\
  《艺术概论》 & 《中外戏剧史》 \\
  《非文学翻译理论与实践》（第二版） & 《翻译论集》 \\

  \rowcolor{agronomy}
  \multicolumn{2}{c}{\cellcolor{agronomy}\textbf{AGRONOMY}} \\
  \midrule
  《农学概论》 & 《森林培育学》 \\
  《植物生理学》 & 《生态学》 \\
  《植物生产学》 & 《植物学》 \\
  《经济林栽培学总论》 & 《园林树木栽培学》 \\
  《森林昆虫学》 & 《森林经理学》 \\
  《土壤学》 & 《植物资源学》 \\
  《水产养殖工程学》 & 《水产养殖学》 \\
  《蔬菜培养学总论》&《兽医病理生理》 \\
  《兽医学概论》& 《作物学》 \\

  \rowcolor{law}
  \multicolumn{2}{c}{\cellcolor{law}\textbf{LAW}} \\
  \midrule
  \textit{Translation Between English And Arabic: A Textbook For Translation Students And Educators} & \textit{Urdu For Children, Book II, 3 Book Set, Part Two: Part 2 Set Of Books} \\
  \textit{Excursions In World Music, Seventh Edition Volume 2} & \textit{Fundamentals, Function, And Form Theory And Analysis Of Tonal Western Art Music} \\
  \textit{News And Journalism In The UK} & \textit{} \\
  《政治科学》 & 《公共管理导论》 \\
  《比较政治学》 & 《公共经济学》 \\
  《世界政治：供选择的菜单》 & 《政治科学研究方法》 \\
  《西方经济学》\\

\end{longtable}

\begin{longtable}{@{}p{4cm}p{8cm}p{2cm}p{1cm}@{}}

\caption{Benchmarks Used } \\
\toprule
\textbf{Benchmark Name} & \textbf{Title} & \textbf{Year} \\ \midrule

\endhead
\bottomrule
\endfoot

LawBench & LawBench: Benchmarking Legal Knowledge of Large Language Models~\citep{fei2023lawbench} & 2023 \\ \midrule
MedMCQA & MedMCQA: A Large-scale Multi-Subject Multi-Choice Dataset for Medical domain Question Answering~\citep{pmlr-v174-pal22a} & 2022 \\ \midrule
MedQA & What Disease does this Patient Have? A Large-scale Open Domain Question Answering Dataset from Medical Exams~\citep{jin2020disease} & 2020 \\ \midrule
MMLU-Pro & MMLU-Pro: A More Robust and Challenging Multi-Task Language Understanding Benchmark~\citep{wang2024mmlupro} & 2024 \\ \midrule
MMLU-CF & MMLU-CF: A Contamination-free Multi-task Language Understanding Benchmark~\citep{zhao2024mmlucfcontaminationfreemultitasklanguage} & 2024 \\ \midrule
ShoppingMMLU & Shopping MMLU: A Massive Multi-Task Online Shopping Benchmark for Large Language Models~\citep{jin2024shopping} & 2024 \\ \midrule
UTMath & UTMath: Math Evaluation with Unit Test via Reasoning-to-Coding Thoughts~\citep{yang2024utmath} & 2024 \\ \midrule
ChatMusician & ChatMusician: Understanding and Generating Music Intrinsically with LLM~\citep{yuan2024chatmusician} & 2024 \\ \midrule
Omni-Math & Omni-MATH: A Universal Olympiad Level Mathematic Benchmark For Large Language Models~\citep{gao2024omnimathuniversalolympiadlevel} & 2024 \\ 
U-MATH & U-MATH: A University-Level Benchmark for Evaluating Mathematical Skills in LLMs~\citep{chernyshev2024umath} & 2024 \\ \midrule
Putnam-AXIOM & Putnam-AXIOM: A Functional and Static Benchmark for Measuring Higher Level Mathematical Reasoning~\citep{fronsdal2024putnamaxiom} & 2024 \\ \midrule
Short-form Factuality & Measuring short-form factuality in large language models~\citep{wei2024measuringshortformfactualitylarge} & 2024 \\ \midrule
Chinese SimpleQA & Chinese SimpleQA: A Chinese Factuality Evaluation for Large Language Models~\citep{he2024chinesesimpleqachinesefactuality} & 2024 \\ \midrule
AIME-AOPS & AIME Problems and Solutions~\citep{aopsAIME} & 2024 \\ \midrule
AIMO Validation AIME & AIMO Validation AIME: Internal Validation Set for AIMO Progress Prize~\citep{aimoValidationAIME} & 2024 \\ \midrule

\end{longtable}

\end{CJK}
\bibliographystyle{unsrt}
\bibliography{main}